\documentclass{article}

\usepackage{latexsym,theorem}
\usepackage{exscale}
\usepackage{xspace} 
\usepackage{pstricks} 
\usepackage{epsfig}
\usepackage[english]{varioref}
\usepackage[english]{babel}
\usepackage{graphics}
\usepackage{array}
\usepackage{alltt}
\usepackage[mathscr]{eucalr}
\usepackage{shadow}
\usepackage{epic}
\usepackage{makeidx}
\usepackage{eepic}
\usepackage{supertabular}
\usepackage{theorem}
\usepackage{url} 
\usepackage{amsmath,amscd,amssymb}
\usepackage{chicago}

{\obeyspaces %
\gdef\sepspaces{\def {\ }}}
\def\EX{%
\parindent=0pt\parskip=0pt\tt\obeylines\obeyspaces\sepspaces}

{\ignorespaces\EX}%
{}

\newcommand{\impact}{\textsl{IMPACT}\index{IMPACT@\textsl{IMPACT}}\xspace}

\newcommand{\corba}{\textsl{CORBA}\index{CORBA@\textsl{CORBA}}\xspace}

\newcommand{\api}{\textsl{API}\index{API@\textsl{API}}\xspace}

\newcommand{\odmg}{\textit{ODMG}\index{ODMG@\textit{ODMG}}\xspace}

\renewcommand{\iff}{\textit{if and only if}\xspace}
\newcommand{\iffdef}{\xspace\textit{if, by definition,}\xspace}

\newcommand{\la}{\ensuremath{\gets}\xspace}

\newcommand{\defeq}{\ensuremath{=_{\mathit{def}}}\xspace}

\newcommand{\false}{\ensuremath{\protect\mathbf{false}}\xspace}
\newcommand{\true}{\ensuremath{\protect\mathbf{true}}\xspace}

\newcommand{\ag}[1]{\ensuremath{\protect\mathscr{#1}\xspace}}

\newcommand{\type}[1]{{\ensuremath{\mathtt{#1}}}\xspace}

\newcommand{\scode}{\ensuremath{\protect\mathcal{S}}\index{S@\ensuremath{\protect\mathcal{S}}}\index{\ensuremath{\protect\mathcal{S}}}\xspace}
\newcommand{\sctype}{\ensuremath{\protect\mathcal{T}}\index{T@\ensuremath{\protect\mathcal{T}}}\index{\ensuremath{\protect\mathcal{T}}}\xspace}     
\newcommand{\scfunct}{\ensuremath{\protect\mathcal{F}}\index{F@\ensuremath{\protect\mathcal{F}}}\index{\ensuremath{\protect\mathcal{F}}}\xspace}    
\newcommand{\sccomp}{\ensuremath{\protect\mathcal{C}}\index{C@\ensuremath{\protect\mathcal{C}}}\index{\ensuremath{\protect\mathcal{C}}}\xspace}    

\newcommand{\var}[1]{\ensuremath{\protect\mathtt{#1}}\xspace} 
\newcommand{\cc}[3]{\ensuremath{{\ag{#1}}\mathit{\,:\,}\mathit{#2}(\type{#3})}\xspace}                                                            

\newcommand{\ccc}{\ensuremath{\chi}\index{\ensuremath{\chi}}\index{chi@\ensuremath{\chi}}}         

\newcommand{\IN}[2]{\ensuremath{\textrm{\normalfont\bfseries{in}}\protect\mathbf{(}\protect\mathtt{#1}\protect\mathbf{,\:}\protect\mathtt{#2}\protect\mathbf{)}}}

\newcommand{\notIN}[2]{\ensuremath{\textrm{\normalfont\bfseries{not\_in}}\protect\mathbf{(}\mathtt{#1}\mathsf{,\:}\mathtt{#2}\protect\mathbf{)}}}

\newcommand{\concur}{\ensuremath{\textbf{conc}}\index{conc@\ensuremath{\textbf{conc}}}\xspace}

\newcommand{\intcons}{\ensuremath{\protect\mathcal{IC}}\index{IC@\ensuremath{\protect\mathcal{IC}}}\index{\ensuremath{\protect\mathcal{IC}}}\xspace}   

\newcommand{\str}[1]{\text{{\EX "}#1{\EX "}}}

\newcommand{\op}{\textsf{Op}\index{\textsf{Op}}\index{Op@\textsf{Op}}\xspace}

\newcommand{\agprog}{\ensuremath{\protect\mathcal{P}}\index{P@\ensuremath{\protect\mathcal{P}}}\index{\ensuremath{\protect\mathcal{P}}}\xspace}

\newcommand{\agstate}{\ensuremath{\protect\mathcal{O}}\index{O@\ensuremath{\protect\mathcal{O}}}\index{\ensuremath{\protect\mathcal{O}}}\xspace}

\newcommand{\aga}{\ensuremath{\mathscr{a}}\xspace}
\newcommand{\agb}{\ensuremath{\mathscr{b}}\xspace}

\newcommand{\actcons}{\ensuremath{\protect\mathcal{AC}}\index{AC@\ensuremath{\protect\mathcal{AC}}}\index{\ensuremath{\protect\mathcal{AC}}}\xspace}

\newcommand{\act}[1]{\ensuremath{\mathit{#1}\xspace}}   

\newcommand{\pre}[1]{\ensuremath{\mathit{Pre}(#1)}\index{Pre@\ensuremath{\protect\mathit{Pre}}}\xspace} 
\newcommand{\add}[1]{\ensuremath{\mathit{Add}(#1)}\index{Add@\ensuremath{\protect\mathit{Add}}}\xspace} 
\newcommand{\del}[1]{\ensuremath{\mathit{Del}(#1)}\index{Del@\ensuremath{\protect\mathit{Del}}}\xspace} 

\newcommand{\App}{\ensuremath{\textbf{App}_{\agprog,\agstate_\scode}(S)}\index{\ensuremath{\textbf{App}_{\protect\mathcal{P},\protect\mathcal{O}_{\protect\mathcal{S}}}(S)}}\index{App@\ensuremath{\textbf{App}_{\protect\mathcal{P},\protect\mathcal{O}_{\protect\mathcal{S}}}(S)}}\xspace}

\newcommand{\vecx}{\ensuremath{{\vec X}}}

\newcommand{\bfF}{\ensuremath{\protect\mathbf{F}}\index{\ensuremath{\protect\mathbf{F}(\cdot)}}\index{F@\ensuremath{\protect\mathbf{F}(\cdot)}}\xspace}
\newcommand{\bfP}{\ensuremath{\protect\mathbf{P}}\index{\ensuremath{\protect\mathbf{P}(\cdot)}}\index{P@\ensuremath{\protect\mathbf{P}(\cdot)}}\xspace}
\newcommand{\bfW}{\ensuremath{\protect\mathbf{W}}\index{\ensuremath{\protect\mathbf{W}(\cdot)}}\index{W@\ensuremath{\protect\mathbf{W}(\cdot)}}\space}
\newcommand{\bfO}{\ensuremath{\protect\mathbf{O}}\index{\ensuremath{\protect\mathbf{O}(\cdot)}}\index{O@\ensuremath{\protect\mathbf{O}(\cdot)}}\xspace}
\newcommand{\bfDo}{\ensuremath{\protect\mathbf{Do\,}}\index{\ensuremath{\protect\mathbf{Do}(\cdot)}}\index{Do@\ensuremath{\protect\mathbf{Do}(\cdot)}}\xspace}

\newcommand{\lfp}{\ensuremath{{\mathit{lfp}}}\xspace}

\newcommand{\dcl}[2]{\ensuremath{{\protect\mathbf{D}\textbf{-}\protect\mathbf{Cl}}_{#2}({#1})}\index{\ensuremath{{\protect\mathbf{D}\textbf{-}\protect\mathbf{Cl}}_{\cdot}(\cdot)}}\index{DCL@\ensuremath{{\protect\mathbf{D}\textbf{-}\protect\mathbf{Cl}}_{\cdot}(\cdot)}}\xspace}
\newcommand{\acl}[2]{\ensuremath{{\protect\mathbf{A}\textbf{-}\protect\mathbf{Cl}}_{#2}({#1})}\index{\ensuremath{{\protect\mathbf{A}\textbf{-}\protect\mathbf{Cl}}_{\cdot}(\cdot)}}\index{ACL@\ensuremath{{\protect\mathbf{A}\textbf{-}\protect\mathbf{Cl}}_{\cdot}(\cdot)}}\xspace}

\newcommand{\app}[2]{\ensuremath{\protect\mathbf{App}_{#1}({#2})}\index{\ensuremath{\protect\mathbf{App}_{\cdot}({\cdot})}}\index{App@\ensuremath{\protect\mathbf{App}_{\cdot}({\cdot})}}\xspace}

\newcommand{\papp}[2]{\ensuremath{\mathrm{p-}\protect\mathbf{App}_{#1}({#2})}\index{\ensuremath{\protect\mathbf{App}_{\cdot}({\cdot})}}\index{App@\ensuremath{\protect\mathbf{App}_{\cdot}({\cdot})}}\xspace}

\newcommand{\prd}{\ensuremath{\wp}\index{\ensuremath{\wp}!as probability distribution}\index{p!\ensuremath{\wp}!as probability distribution}\xspace}
\newcommand{\rv}{\ensuremath{\protect\mathbf{RV}}\index{\ensuremath{\protect\mathbf{RV}}}\index{RV@\ensuremath{\protect\mathbf{RV}}}\xspace}

\newcommand{\probcc}[3]{\ensuremath{{\ag{#1}}\mathit{\,:_{\mathbf{RV}}\,}\mathit{#2}(\type{#3})}\xspace} 

\newcommand{\prmodels}[3]{\ensuremath{\models^{[#1,#2]}_{#3}}\xspace}
\newcommand{\statemodels}[2]{\ensuremath{\models^{[#1,#2]}}\xspace}

\newcommand{\pap}{\textsf{pap}\index{\protect\textsf{pap}}\index{pap@\protect\textsf{pap}}\xspace}
\newcommand{\paps}{\textsf{pap}s\xspace}
\newcommand{\cstrat}{\ensuremath{\otimes}\index{\ensuremath{\otimes}}\index{disj@\ensuremath{\otimes}}\xspace}

\newcommand{\annlang}{\ensuremath{\protect\mathbf{L}^{\mathit{ann}}}\index{\ensuremath{\protect\mathbf{L}^{\protect\mathit{ann}}}}\index{Lann@\ensuremath{\protect\mathbf{L}^{\protect\mathit{ann}}}}\xspace}

\newcommand{\ai}{\ensuremath{\mathsf{ai}}\index{\ensuremath{\protect\mathsf{ai}}}\index{ai@\ensuremath{\protect\mathsf{ai}}}\xspace}
\newcommand{\ann}[2]{\ensuremath{\protect\mathrm{[}#1,#2\protect\mathrm{]}}\xspace}
\newcommand{\accc}[3]{\ensuremath{#1:\langle #2,#3\rangle}\xspace}
\newcommand{\cacc}{\ensuremath{\Gamma}\index{\ensuremath{\Gamma}}\index{Gamma@\ensuremath{\Gamma}}\xspace}
\newcommand{\pagprog}{\ensuremath{\protect\mathcal{PP}}\index{\ensuremath{\protect\mathcal{PP}}}\index{PP@\ensuremath{\protect\mathcal{PP}}}\xspace}

\newcommand{\pross}{\ensuremath{\protect{\mathcal{P}S}}\index{PS@\ensuremath{\protect\mathcal{P}S}}\index{\ensuremath{\protect\mathcal{P}S}}\xspace}

\newcommand{\PApp}{\ensuremath{\textbf{App}_{\pagprog,\agstate_\scode}(\protect\mathcal{P}S)}\index{\ensuremath{\textbf{App}_{\protect\mathcal{PP},\protect\mathcal{O}_{\protect\mathcal{S}}}(\protect\mathcal{P}S)}}\xspace}

\newcommand{\nop}[1]{{}}

\newcommand{\TPP}{\ensuremath{\textbf{T}_{\pagprog,\agstate_\scode}}\index{\ensuremath{\textbf{T}_{\protect\mathcal{PP},\protect\mathcal{O}_{\protect\mathcal{S}}}}}\index{T@\ensuremath{\textbf{T}_{\protect\mathcal{PP},\protect\mathcal{O}_{\protect\mathcal{S}}}}}\xspace}

\newcommand{\SPP}{\ensuremath{\textbf{S}_{\pagprog,\agstate_\scode}}\index{\ensuremath{\textbf{S}_{\protect\mathcal{PP},\protect\mathcal{O}_{\protect\mathcal{S}}}}}\index{S@\ensuremath{\textbf{S}_{\protect\mathcal{PP},\protect\mathcal{O}_{\protect\mathcal{S}}}}}\xspace}
\newcommand{\pSPP}{\ensuremath{\mathrm{p-}\textbf{S}_{\pagprog,\agstate_\scode}}\index{\ensuremath{\textbf{S}_{\protect\mathcal{PP},\protect\mathcal{O}_{\protect\mathcal{S}}}}}\index{S@\ensuremath{\textbf{S}_{\protect\mathcal{PP},\protect\mathcal{O}_{\protect\mathcal{S}}}}}\xspace}

\newcommand{\REDone}[1]{\ensuremath{\textrm{Red}_{1}(#1)}\index{\ensuremath{\textrm{Red}_{1}(\cdot)}}\index{Red@\ensuremath{\textrm{Red}_{1}(\cdot)}}\xspace}
\newcommand{\REDtwo}[1]{\ensuremath{\textrm{Red}_{2}(#1)}\index{\ensuremath{\textrm{Red}_{2}(\cdot)}}\index{Red@\ensuremath{\textrm{Red}_{2}(\cdot)}}\xspace}
\newcommand{\REDthree}[1]{\ensuremath{\textrm{Red}_{3}(#1)}\index{\ensuremath{\textrm{Red}_{3}(\cdot)}}\index{Red@\ensuremath{\textrm{Red}_{3}(\cdot)}}\xspace}

\newcommand{\pagstate}{\ensuremath{\protect\mathcal{O}^p}\index{O@\ensuremath{\protect\mathcal{O}}}\index{\ensuremath{\protect\mathcal{O}}}\xspace}
\newcommand{\cmpos}[1]{\ensuremath{\mathsf{COS}(#1)}\xspace}
\newcommand{\pks}[1]{\ensuremath{\mathsf{PKS}(#1)}\xspace}
\newcommand{\cpks}[1]{\ensuremath{\mathsf{PKS}(#1)}\xspace}
\newcommand{\cals}{\ensuremath{\mathcal{S}}\xspace}
\newcommand{\evl}[2]{\ensuremath{\mathbf{eval}(#1,#2)}\xspace}
\newcommand{\pksset}{\ensuremath{\mathcal{K}}\xspace}
\newcommand{\pPApp}{\ensuremath{\textbf{p-App}_{\pagprog,\pagstate}(\protect\mathcal{P}S)}}

\newcommand{\hide}[1]{}

\newtheorem{theorem}{Theorem}[section]

\newtheorem{corollary}[theorem]{Corollary}
\newtheorem{definition}[theorem]{Definition}
\newtheorem{proposition}[theorem]{Proposition}
\newtheorem{lemma}[theorem]{Lemma}
\newtheorem{Remark}{Remark}[section]
\newtheorem{Convention}{Convention}[section]
\newtheorem{algorithm}{Algorithm}[section]
\newtheorem{Example}[theorem]{Example}
\newenvironment{proof}{\noindent\textbf{Proof:} }{
\ifhmode\nobreak\qed\par\fi\medskip}
\def\qed{\hspace*{\fill}{\rule[-0.5mm]{1.5mm}{3mm}}}

\newenvironment{example}[0]{\begin{Example}\rm}{\end{Example}}
\newenvironment{remark}[0]{\begin{Remark}\rm}{\end{Remark}}

\begin{document}
  \title{\textbf{Probabilistic Agent Programs}\footnote{Most proofs
      are contained in the appendix.}}
  
  \author{J\"{u}rgen Dix\thanks{This work was carried out when the
      author was visiting the University of Maryland from
      January-October 1999.}\\University of Koblenz, Dept.~of Computer
    Science\\ D-56075 Koblenz, Germany\\ \and  Mirco Nanni\\ University
    of Pisa, Dept.~of Computer Science\\ I-56125 Pisa, Italy\\ \and
  V.S.~Subrahmanian\thanks{This work was supported by the Army
      Research Office under Grants DAAH-04-95-10174, DAAH-04-96-10297,
      DAAG-55-97-10047 and DAAH04-96-1-0398, by the Army Research
      Laboratory under contract number DAAL01-97-K0135 and by an NSF
      Young Investigator award IRI-93-57756.}\\Dept. of CS, University
    of Maryland\\  College Park, MD 20752, USA} 
\maketitle
\begin{abstract}
Agents are small programs that autonomously take actions
based on changes in their environment or ``state.''  Over the
last few years, there have been an increasing number of efforts
to build agents that can interact and/or collaborate with other agents.
In one of these efforts,
  \citeN{eite-etal-99a} have shown how agents may be built on top of 
  legacy code. However, 
  their framework assumes that agent states are completely determined,
  and there is no uncertainty in an agent's state.  Thus, their framework
allows an agent developer to specify how his agents will react when the
agent is 100\% sure about what is true/false in the world state. 
In this paper, we
  propose the concept of a \emph{probabilistic agent program} and show
  how, given an arbitrary program written in any imperative language,
  we may build a declarative ``probabilistic''
agent program on top of it which supports
  decision making in the presence of uncertainty.  We provide two
  alternative semantics for probabilistic 
  agent programs.  We show that the second
  semantics, though more epistemically appealing, is more complex to
  compute.  We provide sound and complete algorithms to compute the
  semantics of \emph{positive} agent programs.
\end{abstract}
\section{Introduction}
Over the last few years, there has been increasing interest in the
area of software agents.  
Such  agents provide a wide variety of services including
identification of interesting newspaper articles, software robots that
perform tasks (and plan) on a user's behalf, content based routers,
agent based telecommunication applications, and solutions to logistics
applications.  $\impact$ (see
\url|http://www.cs.umd.edu/projects/impact/|) is a multinational
project whose aim is to define a formal theory of software agents,
implement (appropriate fragments of) the theory efficiently, and
develop an appropriate suite of applications on top of this
implementation.
An \impact agent manages a set of data
types/structures (including a message box) through a set of application
program interface (API) function calls.  
The state of the agent at a given point in time is a set of objects
belonging to these data types.
Each agent has a set of
integrity constraints that its state must always satisfy.  When an
agent's state changes (due to external events such as receipt of a
message), the agent tries to modify its state so that the
integrity constraints are satisfied.  To do this, it has a suite of
actions, and an \emph{agent program} that specifies the operating
principles (what is permitted, what is forbidden, what is obligatory,
etc., and under what conditions?).  \cite{eite-etal-99a,eite-subr-99}
provides a detailed study of the semantics and complexity of such
agents, \cite{eite-etal-99c} contains compile-time and run-time
algorithms, while \cite{aris-etal-99} focuses on system architecture.

Past work on \impact assumes that all agents reason with a complete and
certain view of the world.  However, in many real world applications,
agents have only a partial, uncertain view of what is true
in the world.  Though an agent may need to reason about uncertainty
for many reasons, in this paper, we will assume that the main 
cause of
uncertainty in an agent is due to its state being uncertain.
For example, when an image processing
agent is asked to identify an enemy vehicle,
it might return the fact that vehicle $v_1$
is a T72 tank (with 60--70\% probability) and a T-80 tank (with 20-45\%
probability). However, this raises several problems, the first
of which is that as an action can only be executed if its precondition
is true in the current state, if the agent doesn't know what the
state is, then it cannot determine which of its actions are
executable, and which are not. Second, even if an action is executable,
the state that results may not be precisely determinable either.
One consequence of all this is that the semantics of agent programs
change significantly when such uncertainties arise.

The main contributions (and organization) of this paper may now
be summed up as follows.

\begin{enumerate}
\item In Section~\ref{c7-sec-prelim}, we present a brief overview of
  agents (without any uncertainty involved) as described in
  \cite{eite-etal-99a}.
\item Then, in Section~\ref{c9-pcc:sec}, we define the concept of a
  probabilistic code call, which is the basic syntactic construct
  through which uncertainty in abstract data types
  manifests itself.
\item In Section~\ref{c9-syn:sec}, we define the syntax of
  probabilistic agent programs.  Specifically, we show that
  probabilistic agent programs allow an agent developer to specify the
  permissions, obligations, forbidden actions, etc. associated with an
  agent depending not only on the probabilities that certain conditions
  hold in the agent's state, but also on the developer's
  assumptions about the relationship between these conditions (e.g.
  the probability that a conjunction holds in a given state depends
  not only on the probabilities of the conjuncts involved, but also
  on the dependencies if any between the conjuncts).
\item In Section~\ref{c9-papsem:sec}, we develop three formal
  semantics for probabilistic agent programs which extend each other
  as well as the semantics for (ordinary, non probabilistic) agent
  programs defined by \citeN{eite-etal-99a}.  We also provide results
  relating these diverse semantics.
\item Then, in Section~\ref{c9-sec:computing}, we develop a sound and
  complete algorithm to compute the semantics defined when only
  positive agent programs are considered. We also show that the
  classical agent programs of \citeN{eite-etal-99a} are a special case
  of our probabilistic programs.
\item In Section~\ref{c9-papstrsem:sec}, we provide an alternative,
  Kripke style semantics for agent programs.  In contrast to the
  previous ``family'' of 
  semantics which assume that an agent's precondition must be
  true with 100\% probability for the agent to execute it, this
  semantics also allows an agent to execute it when it is not sure
  (with 100\% probability) that the action's precondition is true. We
  extend all three semantics of agent programs defined earlier in
  Section~\ref{c9-papsem:sec} to handle these intuitions.
  Unfortunately, as we show in this section, this desire for a ``more
  sophisticated'' sematics comes at a high computational price.
\end{enumerate}

\section{Preliminaries}\label{c7-sec-prelim}
In \impact, each agent $\aga$ is built on top of a body of software
code (built in any programming language) that supports a well defined
application programmer interface (either part of the code itself, or
developed to augment the code).  Hence,
associated with each agent $\aga$ is a body of software code
$\scode^{\aga}$ defined as follows.

\begin{definition}[Software Code]
We may characterize the code on top of which an agent is built as a
triple $\scode\defeq (\sctype_\scode,\scfunct_\scode,\sccomp_\scode)$
where:
\begin{enumerate}
\item $\sctype_\scode$ is the set of all data types managed by
  $\scode$,
\item $\scfunct_\scode$ is a set of predefined functions which makes
  access to the data objects managed by the agent available to
  external processes, and
\item $\sccomp_\scode$ is a set of type composition operations. A type
  composition operator is a partial $n$-ary function $c$ which takes
  types $\tau_1,\ldots,\tau_n$ as input, and yields a type
  $c(\tau_1,\ldots,\tau_n)$ as output. 
  As $c$ is a partial function, $c$ may only be
  defined for certain arguments $\tau_1,\ldots,\tau_n$, i.e., $c$ is not
  necessarily applicable on arbitrary types.
\end{enumerate}
\end{definition}
When $\aga$ is clear from context, we will often drop the superscript $\aga$.
Intuitively,
$\sctype_\scode$ is the set of all data types managed by $\aga$,
$\scfunct_\scode$ is the set of all
function calls supported by $\scode$'s application
programmer interface (\api). $\sccomp_\scode$ is the set of ways of
creating new data types from existing data types.  This
characterization of a piece of software code is
widely used (cf. the \textit{Object Data
  Management Group}'s \odmg standard~\cite{catt-93} and the \corba
framework~\cite{sieg-96}).
Each agent also has a message box having a well defined set of
associated code calls that can be invoked by external programs.

\begin{example}[Surveillance Example]
Consider a surveillance application where there are hundreds of
(identical) surveillance agents, and a geographic agent.  The
data types associated with the surveillance and geographic agent
include the standard \texttt{int,bool,real,string,file} data types,
plus those shown below:

\begin{center}
\begin{tabular}{cc}
\textbf{Surveillance Agent} & \textbf{Geographic Agent}\\
\begin{tabular}[t]{|l|}\hline 
\textbf{image:record of}\\
$\:\:$ imageid:file;\\
$\:\:$ day:date;\\
$\:\:$ time:int;\\
$\:\:$ location:string\\
imagedb: \emph{setof} image;\\ \hline
\end{tabular} &
\begin{tabular}[t]{|l|}\hline 
map:$\uparrow$ quadtree;\\
\textbf{quadtree:record of}\\
$\:\:$  place:string;\\
$\:\:$  xcoord:int;\\
$\:\:$  ycoord:int;\\
$\:\:$  pop:int\\
$\:\:$  nw,ne,sw,se:$\uparrow$ quadtree\\
\hline
\end{tabular} \end{tabular} 
\end{center}
A third agent may well merge information from these two
agents, tracking a sequence of surveillance events.

The $\ag{surv}$ agent may support a function
$\cc{surv}{identify}{}$ which takes as input, an image, and
returns as output, the set of all 
identified vehicles in it. It may also support a function called
$\cc{surv}{turret}{}$ that takes as input, a vehicle id, and
returns as output, the type of gun-turret it has.  Likewise, the $\ag{geo}$
agent may support a function $\cc{geo}{getplnode}{}$ which
takes as input a map and the name of a place and returns the set
of all nodes with that name as the place-field, a function
$\cc{geo}{getxynode}{}$ which
takes as input a map and the coordinates of a place and returns the set
of all nodes with that coordinate as the node, a function
called $\cc{geo}{range}{}$ that takes as input a map, an $x,y$
coordinate pair, and a distance $r$ and returns as output,
the set of all nodes in the map (quadtree) that are within
$r$ units of location $(x,y)$.

\emph{Throughout this paper, we will expand on this simple
example and use it to illustrate and motivate the various
definitions in the paper.}
\end{example}

\begin{definition}[State of an Agent]\label{c7-defstate}
The state of an agent at any given point $t$ in time, denoted
$\agstate_\scode(t)$, consists of the
set of all instantiated data objects of types contained in
$\sctype_\scode^{\aga}$.
\end{definition}
An agent's state may change because it took an action, or because it
received a message.
Throughout this paper we will assume that except for appending
messages to an agent \aga's mailbox, another agent \agb cannot
directly change \aga's state.  However, it might do so indirectly by
shipping the other agent a message requesting a change.

\begin{example}\label{surv2:ex}
  For example, the state of the Geographic Agent may consist of two
  quadtrees (one of which, map1, is shown in Figure~\ref{qt:fig}), and
  the type ``map'' may contain two objects, map1, and map2, pointing
  to these two quadtrees, respectively.  (The figure doesn't show
  population values explicitly. Assume the population values are
  20,000 for Loc1, 28,000 for Loc2, 15,000 for Loc3, and 40,000 for
  Loc4, and 8000 for Loc5.)

\begin{center}
\begin{figure}
\epsfig{file=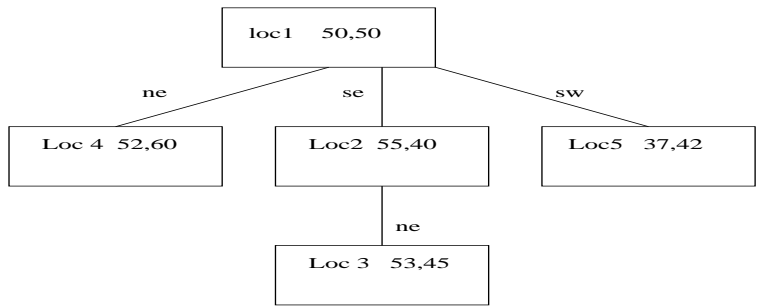,width=10cm,height=4cm}
\caption{Example quadtree for Surveillance Application.}
\label{qt:fig}   
\end{figure}
\end{center}

\end{example}

Queries and/or conditions may  be evaluated w.r.t.  an agent state
using the notion of a code call atom and a code call condition
defined below.

\begin{definition}[Code Call/Code Call Atom]
If $\scode$ is the name of a software package, $f$ is a function
defined in this package, and $(d_1,\ldots ,d_n)$ is a tuple of
arguments of the input type of $f$, then
$\cc{\scode}{f}{d_1,\ldots ,d_n}$ is called a \emph{code call.}

If $\var{cc}$ is a code call, and $\var{X}$ is either a variable
symbol or an object of the output type of $\var{cc}$, then
$\IN{X}{cc}$ is called a \emph{code call atom.}
\end{definition}
For instance, in the Surveillance example,
$\cc{geo}{getplnode}{map1,\str{Loc1}}$ returns
the set containing just the single
node referring to Loc1 in Figure~\ref{qt:fig}.
Likewise, the code call $\cc{geo}{range}{map1,55,50,11}$
returns the set containing the nodes labeled Loc1     
and Loc2.

\begin{definition}[Code Call Condition]
A \emph{code call condition} $\ccc$ is defined as follows:
\begin{enumerate}
\item Every code call atom is a code call condition.
\item If $\var{s,t}$ are either variables or objects, then
  $\var{s=t}$ is a code call condition.
\item If $\var{s,t}$ are either integers/real valued objects, or
  are variables over the integers/reals, then $\mathtt{s < t, s >
    t,s\geq t,s\leq t}$ are code call conditions.
\item If $\chi_1,\chi_2$ are code call conditions, then $\chi_1\,\&\,
  \chi_2$ is a code call condition.
\end{enumerate}
A code call condition satisfying any of the first three criteria
above is an \emph{atomic} code call condition.
\end{definition}
An example code call condition is shown below.
\begin{example}
$\IN{X}{\cc{geo}{range}{map1,55,50,11}}\,\&\, X.pop > 25,000$ is
a code call condition that is satisfied by only one node in 
map1, viz. the Loc2  node.
\end{example}
Each agent has an associated set of \textbf{integrity constraints}---only
states that satisfy these constraints are considered to be
\emph{valid} or \emph{legal} states.  
An integrity constraint is an implication whose consequent is a
code call atom, and whose antecedent is a code call condition.
Appendix A contains a detailed definition.

Each agent has an action-base describing
various actions that the agent is capable of executing.  Actions
change the state of the agent and perhaps the state of other agents'
\type{msgboxes}. 
As in classical AI, all actions have an associated precondition (a code
call condition that the agent state must satisfy for the action to
be executable) and an add/delete list.  Appendix A contains detailed
definitions from \cite{eite-etal-99a}.

For instance, the $\ag{geo}$ agent may have
an \act{insert} action that adds a node to the map.  Likewise,
the \ag{surv} agent may also have an
\act{insert} action  which inserts
a new image  into the image database.  Both these agents also have
an action that sends a message.

Each agent has an associated ``notion of concurrency,'' $\concur$, which
a set of actions and an agent state as input, and produces as
output, a single action that reflects the combination of all the
input actions.  \cite{eite-etal-99a} provides examples of three different
notions of concurrency. 
We will sometimes abuse
notation write $\concur(S,\agstate)$ to denote the 
new state obtained by concurrently executing the actions in $S$ in
state $\agstate$.

Each agent has an associated set
of \emph{action constraints} that define the circumstances under which
certain actions may be concurrently executed.  As at any given point
$t$ in time, many sets of actions may be concurrently executable, each
agent has an \emph{Agent Program} that determines what actions the
agent can take, what actions the agent cannot take, and what actions
the agent must take.
Agent programs are defined in terms of status atoms defined below.

\begin{definition}[Status Atom/Status Set]\label{c7-def:action-status-atom}
  If $\alpha({\vec t})$ is an action, and $Op\in\{
  \bfP,\bfF,\bfW,\bfDo,\bfO\}$, then $Op\alpha({\vec t})$ is called a
  \emph{status atom}.
If $A$ is an action status atom, then $A,\lnot A$ are called
\emph{status literals.}
A \emph{status set} is a finite set of ground status
  atoms.
\end{definition}
Intuitively, $\bfP\alpha$ means $\alpha$ is permitted, $\bfF\alpha$ means
$\alpha$ is forbidden, $\bfDo\alpha$ means $\alpha$ is
actually done, and $\bfW\alpha$ means that the obligation to perform $\alpha$
is waived.

\begin{definition}[Agent Program]
An agent program $\agprog$ is a finite set of rules of the form
\begin{eqnarray*}
A & \gets & \chi\,\&\, L_1\,\&\ldots \&\,  L_n
\end{eqnarray*}
where $\chi$ is a code call condition and $L_1,\ldots ,L_n$ are status
literals.
\end{definition}
The semantics of agent programs are well described in
\cite{eite-etal-99a,eite-subr-99}---due to space reasons, we do not
explicitly recapitulate them here, though Appendix~\ref{app-def} contains
a brief overview of the semantics. 

\section{Probabilistic Code Calls}
\label{c9-pcc:sec}
Consider a code call of the form $\cc{d}{f}{args}$.  This code call
returns a set of objects.  If an object $o$ is returned by such a
code call, then this means that $o$ is \emph{definitely} in the result
of evaluating $\cc{d}{f}{args}$.   However, there are many cases,
particularly in applications involving reasoning about knowledge,
where a code call may need to return an ``uncertain'' answer.   In our
our surveillance example, $\cc{surv}{identify}{image1}$ tries to identify
all objects in a given image---however, it is well known that image
identification is an uncertain task. Some objects may be identified with
100\% certainty, while in other cases, it may only be possible to say
it is either a T72 tank with 40--50\% probability, or a T80 tank with
(50-60\%) probability.

\begin{definition}[Random Variable of Type $\tau$]
  A \emph{random variable} of type $\tau$ is a finite set $\rv$ of
  objects of type $\tau$, together with a probability
  distribution\index{distribution!probability|textbf} $\prd$ that
  assigns real numbers in the unit interval $[0,1]$ to members of
  $\rv$ such that $\Sigma_{o\in\rv}\prd(o) \leq 1$.
\end{definition}
It is important to note that in classical probability theory
\cite{ross-90}, random variables satisfy the stronger requirement that
$\Sigma_{o\in\rv}\prd(o) = 1$.  However, in many real life situations, a
probability distribution may have missing pieces, which explains why
we have chosen a weaker definition.
However, the classical probability case when $\Sigma_{o\in\rv}\prd(o) = 1$
is an instance of our more general definition.

\begin{definition}[Probabilistic Code Call $\probcc{\aga}{f}{d_1,\ldots
    ,d_n}$] \index{code call!probabilistic|textbf}\mbox{}\\ Suppose 
  $\cc{\aga}{f}{d_1,\ldots ,d_n}$ is a code call whose output type
  is $\tau$.  The \emph{probabilistic code call} associated with
  $\cc{\aga}{f}{d_1,\ldots ,d_n}$, denoted
$\probcc{\aga}{f}{d_1,\ldots ,d_n}$,
returns a set of random variables of type $\tau$ 
when executed on state $\agstate$.
\end{definition}
The following example illustrates the use of probabilistic
code calls.

\begin{example}\label{c9-ex2}
Consider the code call $\cc{surv}{identify}{image1}$.  This code call may return
the following two random variables.   
\[ \langle \{ t72,t80\},\{\langle t72,0.5 \rangle, \langle t80,0.4
\rangle\}\rangle \:\: \mathrm{and}\:\: \langle \{ t60,t84\},\{\langle
t60, 0.3 \rangle, \langle t84, 0.7 \rangle\}\rangle \]
This says that the image processing algorithm has identified two
objects in image1. The first object is either a T72 or a T80 tank
with 50\% and 40\% probability, respectively, while the second
object is either a T60 or a T84 tank with 30\% and 70\% probability
respectively.
\end{example}
Probabilistic code calls and code call conditions look exactly like ordinary
code calls and code call conditions
--- however, as a probabilistic code call returns a set of
\emph{random variables}, probabilistic code
call atoms are true or false with some probability.

\begin{example}\label{c9-ex3}
Consider the probabilistic code call condition
\[ \IN{X}{\probcc{surv}{identify}{image1}}\,\&\,
\IN{A1}{\probcc{surv}{turret}{X}}.\]
This code call condition attempts to find all vehicles in
``image1'' with a gun turret of type A1.  Let us suppose that
the first code call returns just one random variable specifying that
image1 contains one vehicle which is either a T72 (probability 50\%) 
or a T80 tank (probability 40\%).
When this random variable (X) is passed to the second code call,
it returns one random variable with two values---A1 with probability
30\% and A2 with probability 65\%.  What is the probability that
the code call condition above is satisfied by a particular assignment
to $X$?

The answer to this question depends very much upon the knowledge we
have (if any) about the dependencies between the identification of a
tank as a T-72 or a T-80, and the type of gun turret on these. For
instance, if we know that all T72's have A2 type turrets, then the
probability of the conjunct being true when X is a T72 tank is 0.
On the other hand, it may be that the turret identification and the
vehicle identification are independent for T80s---hence, when
X is set to T80, the probability of the conjunct being true is
$0.4\times 0.3 = 0.12$.
\end{example}
Unfortunately, this is not the only problem.
Other problems also arise, as shown in the following example.

\begin{example}\label{c9-ex4}
Suppose we consider a code call \ccc~returning the following
two random variables\index{random variable}.
\begin{eqnarray*}
\rv_1 & = & \langle \{ a,b\}, \prd_1\rangle\\
\rv_2 & = & \langle \{ b,c\}, \prd_2\rangle
\end{eqnarray*}
Suppose $\prd_1(a)=0.9,\prd_1(b)=0.1,\prd_2(b)=0.8,\prd_2(c)=0.1$.
What is the probability that $b$ is in the result of the code
call \ccc?
 
Answering this question is problematic.  The reason is that we are
told that there are at most two objects returned by \ccc.  One of
these objects is either $a$ or $b$, and the other is either $b$ or
$c$.  This leads to four possibilities, depending on which of these is
true.  The situation is further complicated because in some cases,
knowing that the first object is $b$ may preclude the second object
from being $b$---this would occur, for instance, if \ccc~examines
photographs each containing two different people and provides
identifications for each. $a$, $b$ and $c$ may be potential id's of
such people returned by the image processing program.  In such cases,
the same person can never be pictured with himself or herself.
 
Of course, in other cases, there may be no reason to believe that
knowing the value of one of two objects tells us anything about the
value of the second object.  For example if we replace people with
colored cubes (with $a$ denoting amber cubes, $b$ black, and $c$
cyan), there is no reason to believe that two identical black cubes
cannot be pictured next to each other.
\end{example}
One could argue, however, that the above reasoning is incorrect 
because if  two objects are
completely identical, then they must be the same.  This means
that if we have two distinct black cubes, then these two black
cubes must be \emph{distinguishable} from one another via some
property such as their location in the photo, or their $\mathit{Id}$s.
This is Leibniz's well known 
\emph{extensionality principle}.
Hence, we will require the results of a probabilistic code
call to be \emph{coherent} in the following sense.
\begin{definition}[Coherent Probabilistic Code Call]
A probabilistic code call is \emph{coherent}  iff
for all distinct $\langle X_1,\prd_1\rangle , \langle X_2,\prd_2\rangle$,
 $X_1\,\cap\,X_2=\emptyset$.
\end{definition}
Throughout this paper, only coherent
probabilistic code calls are considered.  Thus, the expression
``probabilistic code call'' assumes coherence.

\begin{definition}[Satisfying a  Code Call Atom]
Suppose $\probcc{\aga}{f}{d_1,\ldots ,d_n}$ is a ground probabilistic
  code call and $o$ is an object of the output type of this
  code call w.r.t.~agent state $\agstate$.  Suppose $[\ell,u]$ is a
  closed subinterval of the unit interval $[0,1]$.

\begin{itemize}
\item $o\prmodels{\ell}{u}{\agstate} \IN{X}{\probcc{\aga}{f}{d_1,\ldots
      ,d_n}}$
  \\
  if there is a  $(Y,\prd)$ in the answer returned by evaluating
  $\probcc{\aga}{f}{d_1,\ldots ,d_n}$ w.r.t.~$\agstate$ such that
  $o\in Y$ and $\ell\leq \prd(o)\leq u$.
\item $o\prmodels{\ell}{u}{\agstate} \notIN{X}{\probcc{\aga}{f}{d_1,\ldots
      ,d_n}}$
  \\
  if for all random variables $(Y,\prd)$ returned by evaluating
  $\probcc{\aga}{f}{d_1,\ldots ,d_n}$ 
  w.r.t.~$\agstate$, either $o\notin Y$ or $\prd(o)\notin [\ell,u]$.
\end{itemize}
\end{definition}
Probabilistic code call conditions are defined in exactly the
same way as code call conditions. However, extending the above
definition of ``satisfaction'' to probabilistic code call
conditions is highly problematic because (as shown in Examples~\ref{c9-ex3} and ~\ref{c9-ex4}),
the probability that a conjunction is true
depends not only on the probabilities of the individual conjuncts,
but also on the dependencies between the events denoted by these
conjuncts.  The notion of a \emph{probabilistic conjunction
strategy} defined below captures these different ways of
computing probabilities via an abstract definition.

\begin{definition}[Probabilistic Conjunction Strategy $\mathbf{\otimes}$]\mbox{}\\
A \emph{probabilistic conjunction strategy} is a mapping $\otimes$
which maps a pair of probability intervals to a single probability
interval satisfying the following axioms:

\begin{enumerate}
\item \textbf{Bottomline\index{bottomline}:} $[L_1, U_1] \otimes [L_2, U_2] 
\leq [{\rm min}(L_1,
  L_2), {\rm min}(U_1, U_2)]$ where $[x,y]\leq [x',y']$ if
$x\leq x'$ and $y\leq y'$.
\item \textbf{Ignorance\index{ignorance}:}
$[L_1, U_1] \otimes [L_2, U_2] \subseteq
[{\rm max}(0, L_1 + L_2 - 1), {\rm min}(U_1, U_2)]$.
\item \textbf{Identity\index{identity}:}
When $(e_1 \land e_2)$ is consistent and $[L_2, U_2] = [1, 1]$,
$[L_1, U_1] \otimes [L_2, U_2] = [L_1, U_1]$.
\item \textbf{Annihilator\index{annihilator}:}
$[L_1, U_1] \otimes [0, 0] = [0, 0]$.
\item \emph{Commutativity\index{commutativity}:}
$[L_1, U_1] \otimes [L_2, U_2] = [L_2, U_2] \otimes [L_1, U_1]$.
\item \textbf{Associativity\index{associativity}:}
$([L_1, U_1] \otimes [L_2, U_2]) \otimes [L_3, U_3] =
[L_1, U_1] \otimes ([L_2, U_2] \otimes [L_3, U_3])$.
\item \textbf{Monotonicity\index{monotonicity}:}
$[L_1, U_1] \otimes [L_2, U_2] \leq
[L_1, U_1] \otimes [L_3, U_3]$ if $[L_2, U_2] \leq [L_3, U_3]$.
\end{enumerate}
\end{definition}
Intuitively, in the above definition, $[L_1,U_1],[L_2,U_2]$
are intervals in which the probability of events $e_1,e_2$ are
known to lie, and $[L_1, U_1] \otimes [L_2, U_2]$ returns a
probability range for the co-occurrence of both these events.
The Bottomline axiom says that the probability of the conjunct
is smaller than the probabilities of the individual events.
When we know nothing about the relationship between the events $e_1,e_2$,
\citeN{bool-854} has shown that the probability of the conjunction
must lie in the interval 
$[{\rm max}(0, L_1 + L_2 - 1), {\rm min}(U_1, U_2)]$.  This is what
is stated in the Ignorance axiom. The identity and annihilator
axioms specify what happens when one of the events is deterministic
(i.e. not probabilistic).   The axioms of commutativity and associativity
are self explanatory.  The monotonicity axiom says that if we sharpen
the probability range of one of the two events, then the probability
range of the conjunctive event is also sharpened.

The concept of a conjunction strategy is very general, and has as
special cases, the following well known ways of combining probabilities.

\begin{enumerate}
\item When we do not know the dependencies between $e_1,e_2$, we may use
the conjunction strategy $\otimes_{ig}$ defined as
 $([L_1, U_1] \otimes_{ig} [L_2, U_2]) \equiv [{\rm max}(0,
  L_1 + L_2 - 1), {\rm min}(U_1, U_2)]$.
\item When $e_1,e_2$ have maximal overlap, use the positive correlation
conjunctive strategy $\otimes_{pc}$ defined as
$([L_1, U_1] \otimes_{pc}
  [L_2, U_2]) \equiv [{\rm min}(L_1, L_2), {\rm min}(U_1, U_2)]$.
\item When $e_1,e_2$ have minimal overlap, use the negative correlation 
conjunctive strategy $\otimes_{nc}$ defined as 
$([L_1, U_1] \otimes_{nc} 
  [L_2, U_2]) \equiv 
[{\rm max}(0, L_1 +
  L_2 - 1), {\rm max}(0, U_1 + U_2 - 1)]$.
\item When the two events occur independently, use the 
independence conjunction strategy
$([L_1, U_1] \otimes_{in} [L_2, U_2]) = [L_1 \cdot L_2, U_1 \cdot U_2]$.
\end{enumerate}

\section{Probabilistic Agent Programs:  Syntax}
\label{c9-syn:sec}
We are now ready to define the syntax of
a probabilistic agent program   ($\pap$ for short).  This syntax
builds upon the well studied
\emph{annotated logic paradigm} proposed
by\cite{subr-87}, and later studied extensively
\cite{kife-subr-92,ng-subr-93,ng-subr-93b}.

\subsection{Annotation Syntax}
We assume the existence of an \emph{annotation language}
$\annlang$---the constant symbols of $\annlang$ are the real numbers
in the unit interval $[0,1]$.  In addition, $\annlang$ contains a
finite set of function symbols, each with an associated arity, and a
(possibly infinite) set of variable symbols, ranging over the unit
interval $[0,1]$.  All function symbols are \emph{pre-interpreted} in
the sense that associated with each function symbol $f$ of arity $k$
is a fixed function from $[0,1]^k$ to $[0,1]$.

\begin{definition}[Annotation
  Item]
We define annotation items inductively as follows:

\begin{itemize}
\item Every constant and every variable of $\annlang$ is an annotation
  item.
\item If $f$ is an annotation function of arity $n$ and $\ai_1,\ldots ,
  \ai_n$ are annotation items, then $f(\ai_1,\ldots ,\ai_n)$ is an
  annotation item.
\end{itemize}
An annotation item is \emph{ground} if no annotation variables occur
in it.
\end{definition}
For instance, $0, 0.9,(V+0.9),(V+0.9)^2$ are all annotation items
if $V$ is a variable in $\annlang$ and ``$+$'', ``$\hat{\mbox{}}$''
are annotation functions of arity $2$.

\begin{definition}[Annotation $\ann{\ai_1}{\ai_2}$]
  If $\ai_1,\ai_2$ are annotation items, then $\ann{\ai_1}{\ai_2}$ is
  an \emph{annotation}.  If $\ai_1,\ai_2$ are both ground, then
  $\ann{\ai_1}{\ai_2}$ is a \emph{ground annotation}.
\end{definition}
For instance, $\ann{0}{0.4},\ann{0.7}{0.9},\ann{0.1}{\frac{V}{2}},
\ann{\frac{V}{4}}{\frac{V}{2}}$ are all annotations.  The annotation
$\ann{0.1}{\frac{V}{2}}$ denotes an interval only when a value in
$[0,1]$ is assigned to the variable $V$.

\begin{definition}[Annotated Code Call Condition
  $\accc{\chi}{\ann{\ai_1}{\ai_2}}{\cstrat}$]
If $\chi$ is a probabilistic code call
  condition, $\cstrat$ is a conjunction strategy, and
  $\ann{\ai_1}{\ai_2}$ is an annotation, then
  $\accc{\chi}{\ann{\ai_1}{\ai_2}}{\cstrat}$ is an \emph{annotated code
    call condition}.  $\accc{\chi}{\ann{\ai_1}{\ai_2}}{\cstrat}$ is
  \emph{ground} if there are no variables in either $\chi$ or in
  $\ann{\ai_1}{\ai_2}$.
\end{definition}
Intuitively, the ground annotated code call condition 
$\accc{\chi}{\ann{\ai_1}{\ai_2}}{\cstrat}$
says that the
probability  of $\chi$ being
true (under conjunction strategy $\cstrat$)
 lies in the interval $\ann{\ai_1}{\ai_2}$.
For example, when $X,A1$ are ground, 
\[ \accc{\IN{X}{\probcc{surv}{identify}{image1}}\,\&\,
\IN{A1}{\probcc{surv}{turret}{X}}}{\ann{0.3}{0.5}}{\otimes_{ig}}\]
is true \iff  the probability that $X$ is identified by the
\ag{surv} agent and that the turret is identified as being of type A1
lies between 30 and 50\% assuming that nothing is known about
the dependencies between turret identifications and identifications
of objects by \ag{surv}.

We are now ready to define the concept of a probabilistic
agent program.

\begin{definition}[Probabilistic Agent Programs \pagprog]
Suppose $\cacc$ is a conjunction of annotated code calls,
and $A,L_1,\ldots ,L_n$ are action status atoms.  Then
\begin{equation}
\label{c9-parule}
A \la \cacc, L_1,\ldots, L_n
\end{equation}
is a \emph{probabilistic action rule}.
For such a rule $r$, we use $B^+_{as}(r)$ to denote the
positive action status atoms in $\{ L_1,\ldots ,L_n\}$,
and $B^-_{as}(r)$ to denote the set of negative action
status liters in  $\{ L_1,\ldots ,L_n\}$.

A \emph{probabilistic agent program} is a finite set of
probabilistic action rules.
\end{definition}
A simple example of a probabilistic agent program is given below.

\begin{example}[Probabilistic Agent Program]
Consider an intelligent sensor agent
 that is performing surveillance tasks.  The
following rules specify a small \pap that such an agent might use.

\begin{eqnarray*}
\bfDo send\_warn(X) & \gets & 
\IN{F}{\cc{surv}{file}{imagedb}}\,\&\,\\
& & \IN{X}{\probcc{surv}{identify}{F}}\,\&\,
\\
& & \IN{A1}{\probcc{surv}{turret}{X}}):\langle \ann{0.7}{1.0},\otimes_{ig}
\rangle\\
& & \lnot\bfF send\_warn(X).
\\
\bfF send\_warn(X) & \gets &
\IN{X}{\probcc{surv}{identify}{F}}\,\&\,\\
& & \IN{L}{\probcc{geo}{getplnode}{X.location}}\,\&\,\\
& & \IN{L}{\probcc{geo}{range}{100,100,20}}.
\end{eqnarray*}
This agent operates according to two very simple rules.
The first rule says that it sends a warning whenever it
identifies an enemy vehicle as having a gun turret of type
A1 with over 70\% probability, as long as sending such
a warning is not forbidden.  The second rule says that
sending a warning is forbidden if the enemy vehicle is
within 20 units of distance from location (100,100).
\end{example}

\section{Probabilistic Agent Programs:  Semantics}
\label{c9-papsem:sec}
We are now ready to define the semantics of \paps.  The semantics
of \paps will be defined via the concept of a \emph{probabilistic
status set} (defined below).

\begin{definition}[Probabilistic Status Set
  \pross]
A \emph{probabilistic status  set} is any set \pross of 
ground action status
    atoms over $\scode$.  For any operator $\op \in \{
    \bfP,\bfDo,\bfF,\bfO,\bfW \}$, we denote by $\op(\pross)$ the set
    $\{ \alpha \mid \op(\alpha) \in \pross\}$.
Similarly, we use $\lnot\pross $ to denote the set
$\{ \lnot A\: |\: A\in\pross\}$.
\end{definition}
It will turn out that given any probabilistic agent program, and
an (uncertain) agent state evaluated using probabilistic code calls,
the meaning of the \pap w.r.t. the state may be defined via a set
of probabilistic status sets that have some desirable properties. These
properties fall into three broad categories:

\begin{enumerate}
\item the probabilistic status set must be ``closed'' under the
rules in the \pap;
\item the probabilistic status set must be deontically consistent
(e.g. it cannot require something to be both permitted and forbidden)
and it must not violate the action constraints;
\item the probabilistic status set must not lead to a new
state that violates the integrity constraints associated with the
agent;
\end{enumerate}

\subsection{Satisfaction of Annotated Formulae}\label{c9-sat:sec}
In this section, we define what it means for an agent state to satisfy
an annotated code call condition.  

\begin{definition}[Satisfying an Annotated Code Call Condition]
Suppose $\agstate$ is an agent state, and 
$\accc{\chi}{\ann{\ai_1}{\ai_2}}{\cstrat}$ is a ground annotated code
call condition.  $\agstate$ is said to \emph{satisfy}
$\accc{\chi}{\ann{\ai_1}{\ai_2}}{\cstrat}$, denoted
$\agstate\statemodels{\ai_1}{\ai_2}\accc{\chi}{\ann{\ai_1}{\ai_2}}{\cstrat}$
iff:
\begin{itemize}
\item  $\chi$ is of the form $o = o$ (where $o$ is an object), or
\item  $\chi$ is of the form $r_1<r_2$, where  $r_1, r_2$ are
real numbers  (or integers) such that $r_1$ is less than $r_2$, or
\item $\chi$ is of the form $\IN{X}{\probcc{\aga}{f}{d_1,\ldots ,d_n}}$ and
$o\prmodels{\ai_1}{\ai_2}{\agstate} \IN{X}{\probcc{\aga}{f}{d_1,\ldots
,d_n}}$, or
\item $\chi$ is of the form $\notIN{o}{\probcc{\aga}{f}{d_1,\ldots ,d_n}}$
and the following holds
$o\prmodels{\ai_1}{\ai_2}{\agstate} \notIN{X}{\probcc{\aga}{f}{d_1,\ldots
,d_n}}$, or
\item $\chi$ is of the form $\chi_1\,\land\, \chi_2$ and
$[\ell_1,u_1],[\ell_2,u_2]$ are the tightest intervals such that
$\agstate \statemodels{\ell_1}{u_1} \chi_1$ and $\agstate
\statemodels{\ell_2}{u_2} \chi_2$ and $[\ai_1,\ai_2]\supseteq [\ell_1,u_1]
\cstrat
[\ell_2,u_2]$.
\end{itemize}

$\agstate$ is said to \emph{satisfy} a non-ground annotated code call
$\accc{\chi}{\ann{\ai_1}{\ai_2}}{\cstrat}$ iff $\agstate$ satisfies all
ground instances of $\accc{\chi}{\ann{\ai_1}{\ai_2}}{\cstrat}.$
\end{definition}

\subsection{Closure and $\PApp$}\label{c9-closure:sec}
We may associate with any \pap $\pagprog$, an operator $\PApp$ which
maps  probabilistic status sets to  probabilistic status sets.

\begin{definition}[Operator \PApp]
\label{c9:def-papp}
Suppose $\pagprog$ is a probabilistic agent program, $\agstate_\scode$
is an agent state, and $\pross$ is a probabilistic status set.  Then
\[\begin{array}{ll} \PApp = \{ \op\,\alpha\:| & \op\,\alpha \text{ is the head
    of
    a ground instance $r$ of a rule}\\
  & \text{in $\pagprog$ satisfying the 4 conditions below} \, \}
\end{array}\]
\begin{enumerate}
\item $B^+_{as}(r) \subseteq \pross$ and $\lnot.B^-_{as}(r) \cap \pross = \emptyset$, and
\item For every annotated code call condition
  $\accc{\chi}{\ann{\ai_1}{\ai_2}}{\cstrat}$ in the body of $r$, it is
  the case that $\agstate_\scode\statemodels{\ai_1}{\ai_2}
  \accc{\chi}{\ann{\ai_1}{\ai_2}}{\cstrat}$ and
\item if $\op\in \{ \bfP, \bfO, \bfDo\}$, then
  $\agstate_\scode\statemodels{1}{1} Pre(\alpha)$ and
\item for every action status atom of the form $\op\, \beta$ in
  $B^+_{as}(r)$ such that $\op\in \{ \bfP, \bfO, \bfDo\}$,
  $\agstate_\scode\statemodels{1}{1} Pre(\beta)$.
\end{enumerate}
\end{definition}
The first part of this definition says that
for a rule to fire, the action status atoms in its
body must be ``true'' w.r.t. $\pross$.  The second
condition says that annotated code call conditions
in a rule body must be satisfied in the current
object state for the rule to fire.  The third part
is more tricky.  It says that if $\bfO\alpha$
or $\bfDo\alpha$ or $\bfP\alpha$ is in the head of
a rule, then for the rule to fire, the precondition
of the action must be true with 100\% probability.
The final condition is similar w.r.t. to positive
action status atoms in the body.  \emph{Thus, for now,
we are assuming that for an agent to perform an action
(or even be permitted to perform an action),
it must be 100\% sure that the action's precondition
is true} (later in Section~\ref{p-feasible}, we will
provide an alternate, more complex semantics that
does not require this).

\begin{definition}[Closure under Program Rules]
$\pross$ is said to be closed under the rules of \pap
$\pagprog$ in state $\pagstate$ iff
$\PApp\subseteq \pross$.
\end{definition}

\subsection{Deontic/Action Consistency/Closure}
The concept of deontic/action consistency requires that
probabilistic status sets satisfy the agent's action
constraints and commonsense axioms about deontic
modalities.

\begin{definition}[Deontic and Action Consistency]
\label{c9-def:dac}
A probabilistic status set $\pross$ is 
\emph{deontically consistent} 
with respect to an agent state $\agstate$ iff it
satisfies the following rules for any ground action $\alpha$:
    \begin{itemize}
      \item If $\bfO\alpha \in \pross$, then $\bfW\alpha \notin \pross$.
      \item If $\bfP\alpha \in \pross$, then $\bfF\alpha \notin \pross$.
\item If $\bfP\alpha \in \pross$, then
$\agstate\statemodels{1}{1}
Pre(\alpha)$.
\end{itemize}
A  probabilistic status set $\pross$ is \emph{action
consistent} w.r.t. $\agstate$ iff for every action constraint
of the form
\begin{equation}
\label{c9-AC}
\{ \alpha_1(\vecx_1),\ldots ,\alpha_k(\vecx_k)\} \hookleftarrow \chi
\end{equation}
either $\agstate\not\statemodels{1}{1}\chi$ or
$\{ \alpha_1(\vecx_1),\ldots ,\alpha_k(\vecx_k)\}
\not\subseteq \bfDo(\pross)$.
\end{definition}
The following example illustrate the concept of deontic and
action consistency.
\begin{example}
Suppose we have a resource allocation agent having 
two actions --- $send\_A()$ and $send\_B()$ --- each of which
sends a unit of the resource respectively to agents $A$
$B$.
To execute either of them, we need to have at least one unit of
resource, and to execute them together we need at least 2 units:
\[\begin{array}{l}
\pre{send\_A()} = \IN{X}{\cc{allocator}{avail\_rsc}{}} ~\& ~X >
  0. \\
\pre{send\_B()} = \IN{X}{\cc{allocator}{avail\_rsc}{}} ~\& ~X >
  0. \\
\{ send\_to\_A(), send\_to\_B()\} \hookleftarrow
\IN{X}{\cc{allocator}{avail\_rsc}{}} ~\& ~X<2
\end{array}\]
Suppose the agent's current state $\agstate$ is one in which
$avail\_rsc()$ returns
1.  Then 
\[\begin{array}{ll}
\pross = & \{ \bfP send\_to\_A(), \bfDo send\_to\_A(), \\
         & \bfDo send\_to\_B(), \bfO send\_to\_B() \} 
\end{array}\] 

is deontically consistent (there are no $\bfW$ and $\bfF$ atoms at all,
and the action preconditions are true), but not action consistent.
\end{example}
The deontic and action closure of a probabilistic status set
 $\pross$ is defined
in exactly the same way (see appendix, Definition~\vref{c6-def:dcl}) as in the
non-probabilistic case.

\subsection{Probabilistic State Consistency}
The final requirement of a feasible probabilistic status set
ensures that the new state that results after concurrently
executing a set of actions is consistent with the integrity
constraints.

$\agstate$ \emph{satisfies} the integrity
constraint
$ \psi \  \Rightarrow\   \chi$
iff either $\agstate\not\statemodels{1}{1} \psi $ or
$\agstate\statemodels{1}{1} \chi $.

\begin{definition}[Probabilistic State Consistency]
A probabilistic status set $\pross$ is
\emph{probabilistically state consistent} w.r.t.
$\agstate_\scode$ iff the new state,
  $\agstate'_\scode = \concur(\bfDo(\pross),\agstate_\scode)$
  obtained after concurrently executing all actions of the form
  $\bfDo\alpha\in \pross$ satisfies all integrity constraints.
\end{definition}
The following example illustrates the concept of probabilistic     
state consistency.

\begin{example}
Suppose we have a vehicle coordination agent that tracks vehicle
on a road (line), and makes sure that two vehicles do not collide.
Such an agent may have the integrity constraint
\[
\IN{X}{\cc{geo}{getposition}{a}} ~\& ~\IN{Y}{\cc{geo}{getposition}{b}}
\Rightarrow X \not= Y
\]
It may be able to perform an action $move\_forward(a)$:
\[\begin{array}{ll}
\pre{move\_forward(a)} & = \IN{X}{\cc{geo}{getposition}{a}}\\
\del{move\_forward(a)} & = \IN{X}{\cc{geo}{getposition}{a}}\\
\add{move\_forward(a)} & = \IN{X+1}{\cc{geo}{getposition}{a}}
\end{array}\]
In a state $\agstate$ where $\cc{geo}{getposition}{a}$ returns 200,
and $\cc{geo}{getposition}{b}$ returns 201, the status set
\[
\pross = \{ \bfP move\_forward(a), \bfDo move\_forward(a)\}
\]
is not state consistent, as executing $\bfDo(\pross)$ leads to
where both agent $\ag{a}$ and agent $\ag{b}$ are
in position 201, violating the above integrity constraint. 
\end{example}

\subsection{Feasible Probabilistic Status Sets}
The meaning of a \pap (w.r.t. a given state) may be characterized
via those probabilistic status sets that satisfy the conditions
of closure under program rules, deontic/action consistency and
probabilistic state consistency.  Such probabilistic status sets
are said to be \emph{feasible}.

\begin{definition}[Feasible Probabilistic Status Set]\label{c9-feasSS}
Suppose $\pagprog$ is an agent program and
  $\agstate_\scode$ is an agent state. A probabilistic
  status set $\pross$ is  \emph{feasible}
  for $\pagprog$ on $\agstate_\scode$ if the following conditions
  hold:
 
\begin{description}
\item[(\pross1):]  $\PApp \subseteq \pross$ \emph{(closure under the program rules)};
\item[(\pross2):] $\pross$ is deontically
  and action consistent  \emph{(deontic/action consistency)};
\item[(\pross3):] $\pross$ is action closed
  and deontically closed  \emph{(deontic/action closure)};
\item[(\pross4):] $\pross$ is state
  consistent  \emph{(state consistency)}.
\end{description}
\end{definition}
\paps may have zero, one or many feasible status sets,
as seen via the following examples.

\begin{example}\label{ex-feasible-ss}
Consider the following  agent program.
\begin{eqnarray*}
\bfP send\_warn(t80) & \gets & . \\
\bfF send\_warn(t80) & \gets & .
\end{eqnarray*} 
In any agent state $\agstate_\scode$ such that $\agstate_\scode
\statemodels{1}{1} \pre{send\_warn(t80)}$, the above program cannot
have any feasible probabilistic status set $\pross$.
This is because closure under program rules requires that
$\bfP send\_warn(t80),\bfF send\_warn(t80)$ are both in
$\pross$, but this causes $\pross$ to violate deontic consistency.

In contrast, consider the following one-rule program for checking the power
level of surveillance equipment.
\[
\bfO power\_warn() ~\gets ~\IN{X}{\cc{surv}{powerlevel}{}} ~\& ~X<2000.
\]
Suppose $\cc{surv}{powerlevel}{}$ returns 1000 in some
state $\agstate_\scode$, and suppose 
$power\_warn()$ has no preconditions.
If no integrity and action constraints are present, then this \pap
has exactly one feasible status set, viz.
 for $power\_warn()$, and without action
\[ 
\{ \bfO power\_warn(), \bfDo power\_warn(), \bfP power\_warn()\}
\]
in $\agstate_\scode$.

Now let us consider a $\pap$ which says that one of the two agents
$a,b$ must be warned (if it is active).  Furthermore, if $b$ is
to be warned, its regular (non emergency) channel must not be on. 

\begin{eqnarray*}
\bfF open\_ch(b) & \gets & \lnot\bfF open\_ch(b) ~\&
~\bfDo warn\_ag(b). \\
\bfDo warn\_ag(a) & \gets & \IN{a}{\cc{surv}{activeagents}{}}
~\& ~\lnot\bfDo warn\_ag(b). \\
\bfO warn\_ag(b) & \gets & \IN{b}{\cc{surv}{activeagents}{}} ~\&
~\lnot\bfDo warn\_ag(a). 
\end{eqnarray*}
We assume the absence of 
integrity constraints, and preconditions for all actions.
However, the following action constraint is present:
\[
\{warn\_ag(a), warn\_ag(b)\} \hookleftarrow .
\]
If $\cc{surv}{activeagents}{}$
returns $\{a, b\}$ in state $\agstate_\scode$, then
the above program has several feasible status
sets:

\[\begin{array}{l}
\{ \bfF open\_ch(b), \bfO warn\_ag(b), \bfDo warn\_ag(b), \bfP
warn\_ag(b)\} \\
\{ \bfF open\_ch(b), \bfDo warn\_ag(a), \bfP warn\_ag(a) \} \\
\{ \bfDo warn\_ag(a), \bfP warn\_ag(a) \}
\end{array}\]
Notice that no feasible status set contains both
$\bfDo warn\_ag(a)$ and $\bfDo
warn\_ag(b)$.
\end{example}

\subsection{Rational Probabilistic Status Sets}
As seen from the above examples, feasible status sets may
contain action status atoms that are not required for
feasibility.  Rational probabilistic status sets
refine this definition.

\begin{definition}[Groundedness; Rational Probabilistic Status Set]\mbox{}\\
 A  probabilistic status set $\pross$ is \emph{grounded}
if there is no probabilistic status set
  $\pross'\neq \pross$ such that $\pross' \subseteq \pross$ and $\pross'$ 
satisfies
conditions $(\pross1)$--$(\pross3)$ of a feasible probabilistic status set.

$\pross$ is \emph{rational} iff it is feasible and grounded.
\end{definition}

\begin{example}
Consider the last case in Example~\ref{ex-feasible-ss}.
Only two of the listed feasible status sets are rational, viz.
\[\begin{array}{l}
\{ \bfF open\_ch(b), \bfO warn\_ag(b), \bfDo warn\_ag(b), \bfP
warn\_ag(b)\} \:\:\mbox{and}\\
\{ \bfDo warn\_ag(a), \bfP warn\_ag(a) \}
\end{array}\]
\end{example}
 
\subsection{Reasonable Probabilistic  Status Sets}
As we can see from the preceding example, certain
action status atoms may be true in a rational status
set even though there is no rule whose head contains
(or implies) that action status atom.
The concept of a reasonable status set (which is derived
from the well known stable model semantics of logic
programs \cite{gelf-lifs-88}) prevents this.

\begin{definition}[Reasonable Probabilistic Status Set]
\label{reasonable PSS}
Suppose $\pagprog$ is a \pap,  $\agstate_\scode$ is an agent
state, and $\pross$ is a  probabilistic status set.

\begin{enumerate}
\item If $\pagprog$ is a positive (i.e. $B^-_{as}(r)=\emptyset$
for all $r\in\pagprog$), then
  $\pross$ is a \emph{reasonable probabilistic status set} for $\pagprog$ on
  $\agstate_\scode$, \iffdef $\pross$ is a rational probabilistic status set\index{status set!probabilistic} for
  $\pagprog$ on $\agstate_\scode$.
\item Otherwise, the reduct of $\pagprog$ w.r.t.~$\pross$ and $\agstate_\scode$,
  denoted by $red^\pross(\pagprog,\agstate_\scode)$, is the program which
  is obtained from the ground instances of the rules in $\pagprog$
  over $\agstate_\scode$ as follows.
\begin{enumerate}
\item First, remove every rule $r$ such that $B^-_{as}(r) \cap \pross \neq 
\emptyset$;
\item Remove all atoms in $B^-_{as}(r)$ from the remaining rules.
\end{enumerate}
Then $\pross$ is a \emph{reasonable probabilistic status set} for
$\pagprog$ w.r.t.~$\agstate_\scode$, if it is a reasonable
probabilistic status set\index{status set!probabilistic} of the program
$red^\pross(\pagprog,\agstate_\scode)$ with respect to
$\agstate_\scode$. {}
\end{enumerate}
\end{definition}
The following example illustrates the concept of a reasonable
status set.

\begin{example}
Consider again the last case in Example~\ref{ex-feasible-ss}.
Only one of the listed feasible status sets is  reasonable, viz.
\[
\pross = \{ \bfDo warn\_ag(a), \bfP warn\_ag(a) \}
\]
To see why this probabilistic status set is feasible, note that
the reduct of $\pagprog$ w.r.t. $\pross$ is:
\begin{eqnarray*}
\bfDo warn\_ag(a) & \gets & \IN{a}{\cc{surv}{activeagents}{}}.
\end{eqnarray*}
whose (unique) rational status set is obviously $\pross$.
\end{example}

\subsection{Semantical Properties}
In this section, we prove some properties about the different
semantics described above.

\begin{proposition}[Properties of Feasible Status
  Sets]\label{c9-prop:feas}
Let $\pross$ be a feasible probabilistic status set. Then,
\begin{enumerate}
\item If $\bfDo(\alpha) \in \pross$, then $\agstate_\scode \statemodels{1}{1}
  \pre{\alpha}$;
  \item If $\bfP\alpha \notin \pross$, then $\bfDo(\alpha) \notin \pross$;
  \item  If $\bfO\alpha \in \pross$, then $\agstate_\scode \statemodels{1}{1} \pre{\alpha}$;
  \item  If $\bfO\alpha \in \pross$, then $\bfF\alpha \notin \pross$.
\end{enumerate}
\end{proposition}
The following theorem says that reasonable status sets are
rational.

\begin{theorem}[Reasonable Status Sets are Rational]
\label{c9-prop:reasonable-sub-rational}
Let $\pagprog$ be a probabilistic agent program and $\agstate_\scode$
an agent state. Every reasonable probabilistic status set 
of $\pagprog$ on
$\agstate_\scode$ is a rational probabilistic status set
of $\pagprog$ on
$\agstate_\scode$.
\end{theorem}
Given any \pap $\pagprog$ and agent state $\agstate_\scode$,
we may define an operator that maps probabilistic status sets to
probabilistic status sets as follows.
We use then notation
$\dcl{\pross}{}$ to denote the closure of $\pross$ under the
rule $\bfO\alpha\in\pross\Rightarrow\bfP\alpha\in\pross$ and
$\acl{\pross}{}$ to denote the closure of $\pross$ under the 
rules $\bfO\alpha\in\pross\Rightarrow\bfDo\alpha\in\pross$ and 
$\bfDo\alpha\in\pross\Rightarrow\bfO\alpha\in\pross$.

\begin{definition}[$\TPP$ Operator]
\label{c9-tpdef}
Suppose $\pagprog$ is a probabilistic agent program and $\agstate_\scode$ is an
agent state.
Then, for any probabilistic status set $\pross$,
\[\TPP(\pross) = \PApp \cup \dcl{\pross}{} \cup \acl{\pross}{}.\]
\end{definition}

Note that as $\dcl{\pross}{}\subseteq \acl{\pross}{}$, we may equivalently 
write this as
\[
\TPP(\pross) = \PApp \cup \acl{\pross}{}.
\]
The following property of feasible probabilistic status sets is easily seen.
\begin{lemma}[\pross as Prefixpoint of \TPP]\index{prefixpoint}
\label{c9-lem:fixpoint}
Let $\pagprog$ be a probabilistic agent program, $\agstate_\scode$
be any agent state, and $\pross$ be any probabilistic status set.
If $\pross$ satisfies
conditions $(\pross1)$ and $(\pross3)$ of feasibility, then
$\pross$ is pre-fixpoint of $\TPP$, i.e., $\TPP(\pross) \subseteq \pross$.
\end{lemma}
\begin{proof}
Suppose $\op(\alpha)\in \TPP(\pross) = \PApp \cup \acl{\pross}{}$.
Then we have either $\op(\alpha)\in \PApp$ or
$\op(\alpha)\in \acl{\pross}{}$.
By condition (\pross1) defining a feasible probabilistic status set, we know that
$\PApp \subseteq \pross$.
By condition (\pross3), $\pross=\acl{\pross}{}$ and hence, $\acl{\pross}{}\subseteq \pross$.
Therefore, $\TPP(\pross)\subseteq \pross$.
\end{proof}

The following theorem says that in the absence of integrity constraints,
a \pap has a rational probabilistic status
set \iff it has a feasible one.

\begin{theorem}[Existence of Rational Probabilistic Status Sets]
\label{c9-prop:pos-rat-exists}
Let $\pagprog$ be a probabilistic agent program.
If $\intcons = \emptyset$, then $\pagprog$ has a
rational probabilistic status set\index{status set!probabilistic!rational} \iff
$\pagprog$ has a
feasible probabilistic status set.
\end{theorem}

\section{Computing Probabilistic Status Sets of Positive \paps}
\label{c9-sec:computing}
In this section, we present a sound and complete algorithm
to compute the unique reasonable status set of a positive
\pap.  For this purpose, we use a variant of the $\TPP$
operator introduced earlier. This operator, denoted
$\SPP$, is defined as 
\[ \SPP(\pross)=\acl{\app{\pagprog,\agstate_\scode}{\pross}}{}.\]
Computationally, we may compute the operator
$\SPP$ using algorithm~\ref{c9-alg1} below.

\begin{algorithm}[$\SPP$ Computation for Positive \paps]\mbox{}\label{c9-alg1}

\noindent \textbf{Compute-\SPP($\agstate_\scode$: agent state,\
  $\pross$: probabilistic status set)}

\begin{tabbing}
($\star$\  \textbf{Output:}\= a deontically and action consistent set
$\TPP(\pross)$ (if existent) \   \ \= $\star$)\kill
($\star$\  the probabilistic agent program $\pagprog$ is positive; \ \> \> 
$\star$) \\
($\star$\  \textbf{Input:}\>  \ an agent state $\agstate_\scode$, and a prob.
status set $\pross$ \> $\star$) \\
($\star$\  \textbf{Output:}\> \ a deontically and action consistent set
$\SPP(\pross)$ (if existent)\   \> $\star$) \\
($\star$ \> \ or \emph{``no consistent set exists''}\> $\star$) \\[1.5ex]
\qquad\=1.\ \quad \= $X := \pross$; \+\\
2. \textbf{for each} rule  ${r\in \pagprog}$\\
3. \> \textbf{for each} ground instance ${r\theta}$ of $r$\\
4. \> \textbf{if} \= ${r\theta}=\op\,\alpha \la \cacc, L_1,\ldots, L_n$ \textbf{and}\\
  \> \> $\agstate_\scode\models\cacc$ and
        $\{ L_1,\ldots ,L_n\}\subseteq \pross$
  \textbf{and}\\
  \> \> \text{for every atom $\op^{\prime}(\beta) \in \{ L_1,\ldots ,L_n\} \cup
  \{\op(\alpha)\}$}\\
\> \> \text{such that $\op^{\prime}
  \in \{ \bfP, \bfO, \bfDo\}$}: \text{$\agstate_\scode\statemodels{1}{1}
          Pre(\beta)$}\\
5. \>  \textbf{then} \ \=  $X := X\,\cup\acl{\{ \op\,\alpha\}}{}$,\\
6. \>    \>   \textbf{if} \ $X$ \text{ contains } ($\bfO \alpha$ \text{ and }
$\bfW \alpha$) \text{ or }  ($\bfP \alpha$ \text{ and }
$\bfF \alpha$) \\
7. \> \> \textbf{then}  \textbf{Return} \emph{``no consistent set exists''}.\\
8.  \textbf{Return} $X$.\\
  \textbf{end}.
\end{tabbing}
\end{algorithm}
The behavior of Algorithm~\ref{c9-alg1} is illustrated by the following
example.

\begin{example}\label{ex-new-surveillance}
Consider the following program, saying that whenever a (probably)
enemy vehicle $Y$ is detected, 
a warning message about $Y$ is sent to a friendly source and the
agent perfroming the detection is not allowed to move.

\begin{eqnarray*}
\bfO send\_warn(Y) & \gets & 
\IN{F}{\cc{surv}{file}{imagedb}}\,\&\,\\
& & \IN{Y}{\probcc{surv}{identify}{F}}\langle \ann{0.5}{1.0},\otimes_{ig}
\rangle\,\&\,\\
& & \bfO send\_warn(X).
\\
\bfF move() & \gets & \bfDo send\_warn(X).
\\
\bfO send\_warn(X) & \gets & 
\IN{F}{\cc{surv}{file}{imagedb}}\,\&\,\\
& & \IN{X}{\probcc{surv}{identify}{F}}\,\&\,\\
& & \IN{X}{\cc{surv}{enemyvehicles}{}}):\langle \ann{0.5}{1.0},\otimes_{ig}
\rangle.
\end{eqnarray*}
Moreover, assume that in the current state $\agstate_\scode$,
$\cc{surv}{file}{imagedb}$ returns $image1$,
$\cc{surv}{identify}{image1}$ returns the random variables $\langle
\{t80\}, \{\langle t80,0.6\rangle \} \rangle$ and $\langle \{t72\},
\{\langle t72,0.5 \rangle \}
\rangle$, and $\cc{surv}{enemyvehicles}{}$ returns $t80$.

Now we apply Algorithm~\ref{c9-alg1} to compute
$\SPP(\agstate_\scode, \emptyset)$:
Step 1 sets $X=\emptyset$, step 2 selects the first rule, while step 3 
considers all  ground
instances (of the first rule)
whose body's truth is checked in step 4: no instance
satisfies it (because $\pross$ is empty), so nothing happens.
The same result is obtained when step 2 considers the second rule.
Eventually, step 2 considers the third rule, which satisfies the
condition of step 4 with its head instantiated to $\bfO send\_warn(t80)$.
Step 5 inserts $\bfO send\_warn(t80)$, $\bfDo send\_warn(t80)$,
$\bfP send\_warn(t80)$
into $X$. There is no deontic inconsistency, so the check n step
6 fails and then we jump to step 8, which returns the result:
\[
X = \{ ~ \bfO send\_warn(t80), \bfDo send\_warn(t80), \bfP
send\_warn(t80) ~\}.
\]
\end{example}

The operator $\SPP$ may be iteratively applied as follows.

\begin{eqnarray*}
\SPP^0 & = & \emptyset.\\
\SPP^{i+1} & = & \SPP(\SPP^i).\\
\SPP^\omega & = & \bigcup_{i=0}^\infty \SPP^i.
\end{eqnarray*}
The following theorem says that for positive \paps, operator $\SPP$ is
monotonic, continuous, and has a (unique) least fixpoint.

\begin{lemma}[Monotonicity and Continuity of $\SPP$]
\label{c9:th1}
Suppose $\pagprog$ is a positive \pap.  Then the operator $\SPP$ is
monotone and continuous, i.e.
\begin{enumerate}
\item $\pross_1\subseteq \pross_2\Rightarrow \SPP(\pross_1)\subseteq \SPP(\pross_2)$,
\item $\SPP(\bigcup_{i=0}^\infty \pross_0) = \bigcup_{i=0}^\infty \SPP(\pross_i)$ for
any chain $\pross_0 \subseteq \pross_1 \subseteq \pross_2 \subseteq \cdots$ of probabilistic
status sets,
\item $\SPP^\omega$ is a fixpoint of $\SPP$.  Moreover, it is the least
  fixpoint of $\SPP$.
\end{enumerate}
\end{lemma}
\begin{proof}
By the well known Knaster/Tarski theorem, \emph{3.} follows from
\emph{1.} and \emph{2.}.

To show \emph{1.}, let $\pross_1\subseteq \pross_2$.  But then 
\[\app{\pagprog,\agstate_\scode}{\pross_1} \subseteq
\app{\pagprog,\agstate_\scode}{\pross_2},\]
because of the monotonicity of $\app{\pagprog,\agstate_\scode}{}$
 (see~\cite{lloy-84,apt-90}). This implies 
$\acl{\app{\pagprog,\agstate_\scode}{\pross_1}}{}\subseteq
 \acl{\app{\pagprog,\agstate_\scode}{\pross_2}}{}$.

\emph{2.}  follows similarly from the continuity of 
$\app{\pagprog,\agstate_\scode}{}$ and the fact that 
\[\acl{\bigcup_{i=0}^\infty \app{\pagprog,\agstate_\scode}{\pross_i}}{}=\bigcup_{i=0}^\infty \acl{\app{\pagprog,\agstate_\scode}{\pross_i}}{}.\]
\end{proof}
The following example shows the computation of $\SPP^\omega$.

\begin{example}\label{ex-spp-lfp}
Consider the program in Example~\ref{ex-new-surveillance}.
Applying the operator $\SPP$ iteratively, we obtain:
\begin{eqnarray*}
\SPP^0 & = & \emptyset.
\\ 
\SPP^1 & = & \{ \bfO send\_warn(t80), \bfDo
send\_warn(t80), \bfP send\_warn(t80) \} 
\\ 
\SPP^2 & = & \SPP^1 \cup
\{ \bfO send\_warn(t72), \bfDo send\_warn(t72), \\
& & \bfP send\_warn(t72), \bfF move()\} 
\\ 
\SPP^3 & = & \SPP^2 = \SPP^\omega
\end{eqnarray*}
\end{example}

The following results tell us that Lemma~\ref{c9-lem:fixpoint}, which
holds for arbitrary programs, can be strengthened to the case of
positive probabilistic programs.

\begin{theorem}[Rational Probabilistic Status Sets as Least Fixpoints]
\label{c9-theo:pos-lfp}
Suppose $\pagprog$ is a positive \pap, and 
$\agstate_\scode$ is an agent state.  Then:
$\pross$ is a rational probabilistic status set \iff
$\pross = \lfp(\SPP)$ and $\pross$ is a feasible probabilistic status
set.  Recall that $\lfp$ stands for least fixpoint.
\end{theorem}

\begin{corollary}
\label{c9-coroll:pos-unique-rational}
Let $\pagprog$ be a positive probabilistic agent program. Then, on
every agent state $\agstate_\scode$, the rational probabilistic status set
of $\pagprog$ (if one exists) is unique, i.e., if $\pross, \pross'$ are
rational probabilistic status sets for $\pagprog$ on $\agstate_\scode$, then
$\pross=\pross'$.
\end{corollary}
An important corollary of this theorem is that to compute a
reasonable feasible status set  of a \pap, all we need to do it
to compute $\lfp(\SPP)$.  This may be done via Algorithm
{\bf Compute-\lfp} below.

\begin{algorithm}[Reas. Prob. Status Set Computation for
  Positive \paps]\mbox{}\label{c9-alg2}

\noindent \textbf{Compute-\lfp(\TPP): agent state,\
  $\pagprog$: probabilistic agent program)}

\begin{tabbing}
($\star$\  the probabilistic agent program $\pagprog$ is positive; \ \= $\star$) \\
($\star$\  \textbf{Input:}  an agent state $\agstate_\scode$,  and a \pap 
$\pagprog$ \> $\star$) \\
($\star$\  \textbf{Output:}  a  reasonable probabilistic  status set \> $\star$) \\[1.5ex]
\qquad\=1.\ \quad \= change := \true; $X:=\emptyset$;\+\\
2.  \textbf{while} change \textbf{do}\\
3. \> $newX$ = \textbf{Compute-\SPP(X)};\\
4. \> \textbf{if} $newX:=\emph{``no consistent set exists''}$\\
5. \> \textbf{then} \textbf{return} no reasonable  prob. status set exists.\\
6. \> \textbf{if} $X\neq newX$ \textbf{then} $X:=newX$\\
7. \> \textbf{else} change := \false.\\
8.  \textbf{end while};\\
9.  \textbf{if} \=$X$ satisfies all the following conditions\\
10.  \>  $\bullet$\:\: $\bfDo\alpha\in X\Rightarrow \agstate\statemodels{1}{1} Pre(\alpha)$;\\
11.  \> $\bullet$\:\: \=The new state obtained by executing $\concur(\{ \bfDo\alpha
\:|\: \bfDo\alpha\in X\}$\\
12.  \> \> satisfies the integrity constraints;\\
13.  \> $\bullet$\:\: $\{ \bfDo\alpha \:|\: \bfDo\alpha\in X\}$ satisfies the
action constraints.\\
14.  \textbf{then} \textbf{return} $X$\\
15.  \textbf{else} \textbf{return} no reasonable  prob. status set exists.\\
  \textbf{end}.

\end{tabbing}

\end{algorithm}

\begin{theorem}[Polynomial Data Complexity]
\label{c9:th3}
Algorithm {\bf Compute-\lfp} 
has polynomial data-complexity.
\end{theorem}
\begin{proof}
  It is easy to see that the \textbf{while} loop of the algorithm can
  be executed in polynomial time (data-complexity).  Checking if $X$
  satisfies the three conditions at the end of the algorithm are each
  polynomial time checks (assuming the existence of a polynomial
  oracle to compute code call conditions).
\end{proof}

The following example walks the reader through the detailed working of
this algorithm on the motivating example \pap introduced earlier on
in this paper.

\begin{example}
We apply the above algorithm to the program (and agent state) of
Example~\ref{ex-new-surveillance}:
\begin{itemize}
\item Step 1 initialize X to $\emptyset$, i.e. to $\SPP^0$, while steps 2--8
  iteratively apply the procedure which implements the operator
  $\SPP$.
\item At the first iteration, in step 3 $newX$ becomes $\SPP^1$ (shown
  in Example~\ref{ex-spp-lfp}); since there are no deontic
  inconsistencies, the test in step 4 fails, and then we skip to step
  6 which will assign $X := newX$, and then the cycle starts again.
\item At the second iteration $newX$ becomes $\SPP^2$, then it is
  still inconsistency-free and different from $X$, so that we go on
  for another iteration.
\item The third iteration is also the last one, since $\SPP^3 =
  \SPP^2$, and so we skip to the tests in steps 10--13.
\item In our example, the preconditions of all actions are empty and
  then satisfied, there are not integrity constraint and then $X$ is
  trivially integrity consistant, and eventually, there are no action
  constraints and then $X$ is also action consistent.  $X$ is then
  returned as the (unique) reasonable status set of the program.
\end{itemize}
  
\end{example}

\subsection{Agent Programs are  Probabilistic Agent Programs}
In this section, we show that the concept of an agent program defined
by \citeN{eite-etal-99a} is a special case of the framework defined
here.  Hence, \paps generalize the Eiter et.~al.~semantics.
Furthermore, algorithm {\bf Compute-\lfp} may be used to compute
reasonable status sets of positive agent programs.  First, we show how
agent programs may be captured as \paps.

\begin{definition}
  Let \pagprog be a probabilistic agent program, $\pross$ a
  probabilistic status set and \agstate a probabilistic agent state.
  Assume further that each random variable contains exactly one object
  with probability 1.  Then we can define the following mappings:

\begin{description}
\item[\REDone{\cdot},] which maps every probabilistic code call of the
  form $ \langle\{{o}\}, 1\rangle$ to $o$: \[\REDone{\langle\{{o}_{\rv}\}, 1 \rangle}=
  o.\]
\item[\REDtwo{\cdot},] which maps annotated code call conditions to code
  call conditions by simply removing the annotations and the
  conjunction strategy:
  \[\REDtwo{\accc{\chi}{\ann{\ai_1}{\ai_2}}{\cstrat}}=\chi.\]
  We can easily extend $\REDtwo{\cdot}$ to a mapping from arbitrary
  conjunctions of annotated code calls to conjunctions of code calls.
\item[\REDthree{\cdot},] which maps every probabilistic agent program
  to a non-probabilistic agent program: it clearly suffices to define
  $\REDthree{\protect\cdot}$ on probabilistic agent rules. This is done as
  follows
\[\REDthree{A \la \cacc, L_1,\ldots, L_n}= A \la \REDtwo{\cacc}, L_1,\ldots, L_n.\]
\end{description}
\end{definition}
Under the above assumptions, the following theorem
holds.

\begin{theorem}[Semantics of Agent Programs as an Instance of \paps]
Suppose all random variables have the
  form
\[\langle\{\text{object}_{\rv}\}, 1\rangle. \]
Then:
($\accc{\chi}{\ann{\ai_1}{\ai_2}}{\cstrat}$ is
a ground annotated code call condition, $\agstate_\scode$ an agent state)
\begin{description}
\item[Satisfaction:] the satisfaction relations coincide, i.e.
\[\agstate\statemodels{\ai_1}{\ai_2}\accc{\chi}{\ann{\ai_1}{\ai_2}}{\cstrat}
\
\iff \ \agstate \models \REDtwo{\accc{\chi}{\ann{\ai_1}{\ai_2}}{\cstrat}}. \]
\item[App-Operators:] the App-Operators coincide, i.e.
\[\textbf{App}_{\REDthree{\pagprog},\agstate_\scode}(\pross)=
\textbf{App}_{\pagprog,\agstate_\scode}(\pross),\]
where the operator on the left hand side is the one introduced in
Definition~\vref{c7-def:appold}.
\item[Feasibility:] Feasible probabilistic status sets coincide with
  feasible status sets under our reductions, i.e.~ \pross is a
  feasible probabilistic status set\index{status set!probabilistic} w.r.t.
$\pagprog$ \iff \pross is a
  feasible status set w.r.t.~$\REDthree{\pagprog}$.
\item[Rational:]  Rational probabilistic status sets coincide with  rational status sets under our reductions, i.e.~ \pross is a
  rational probabilistic status set\index{status set!probabilistic} w.r.t.
$\pagprog$  \iff \pross is a
  rational status set w.r.t.~$\REDthree{\pagprog}$.
\item[Reasonable:]  Reasonable probabilistic status sets coincide with
  reasonable status sets under our reductions, i.e.~ \pross is a
  reasonable probabilistic status set\index{status set!probabilistic} w.r.t.
$\pagprog$  \iff \pross is a
  reasonable status set w.r.t.~$\REDthree{\pagprog}$.
\item[Computation of Status Sets:] The computations of probabilistic
  status sets given in Algorithms~\vref{c9-alg1} and~\vref{c9-alg2}
  for a \pap \pagprog reduce to the computation of status sets for
  \REDthree{\pagprog}.
\end{description}
\end{theorem}
\begin{proof}
The first two statements are immediate. Feasibility requires
checking conditions ($\pross1$)--($\pross4$), and therefore reduces to the
first two statements. Rational and reasonable status sets are
handled in a completely analogous manner.

That our algorithms reduce to the non-probabilistic case under our
general assumption is trivial: the difference is only the satisfaction 
relation $\statemodels{1}{1}$ which, by the first statement, coincides 
with $\models$.
\end{proof}

\section{Probabilistic Agent Programs:  Kripke Semantics}
\label{c9-papstrsem:sec}
The definition of a feasible status set given in
Section~\ref{c9-papsem:sec} makes several simplifying assumptions.
First (see Definition~\ref{c9:def-papp}), it assumes that an action
can be executed only if its precondition is believed by the agent to
be true in the agent state with probability $1$.  Second (see
Definition~\ref{c9-def:dac}), every action that is permitted must also
have a precondition that is believed to be true with probability $1$.
In this section, we propose a Kripke-style semantics for agent
programs that removes these conditions.

To do this, we will  start by noting that in a probabilistic
state $\pagstate$, the agent returns a set of random variables for
each code call.  Every probabilistic state implicitly determines
a set of (ordinary) states that are ``compatible'' with it.
We use the notation $\evl{\cc{\aga}{f}{d_1,\ldots ,d_n}}{\agstate}$
to denote the result of evaluating the code call
$\cc{\aga}{f}{d_1,\ldots ,d_n}$ w.r.t. the state $\agstate$.

\begin{definition}[Compatibility of State w.r.t. a Probabilistic
  State] 
Let $\pagstate$ be a probabilistic agent state.  An (ordinary)
agent state $\agstate$ is said to be \emph{compatible} with $\pagstate$
iff for every ground code call $\cc{\aga}{f}{d_1,\ldots ,d_n}$, it is
the case that for
every object $o\in\evl{\cc{\aga}{f}{d_1,\ldots ,d_n}}{\agstate}$,
there exists a random variable
$(X,\wp)\in
\evl{\cc{\aga}{f}{d_1,\ldots ,d_n}}{\pagstate}$ such that $o\in X$ and
$\wp(o) > 0$, and there is no other object $o'\in X$ such that
$o'\in\evl{\cc{\aga}{f}{d_1,\ldots ,d_n}}{\agstate}$.

\end{definition}
The following example illustrates this concept.

\begin{Example}
\label{example_compatible_state}
Consider a probabilistic agent state $\pagstate$ with  only
two code calls $\cc{surv}{identify}{image1}$ and
$\cc{surv}{location}{image1}$, which respectively return the random
variables \[\langle \{t80, t72, t70\},\{\langle t80, 0.3 \rangle, \langle
t72, 0.7 \rangle, \langle t70, 0.0 \rangle \} \rangle\]
 and
$\langle \{Loc2\},\{\langle Loc2, 0.8\rangle\} \rangle $. 
The agent states compatible w.r.t. $\pagstate$ are described in the following table:
\begin{center}
\begin{tabular}{r|cc}
\textbf{State} & \textbf{Vehicle} & \textbf{Location} \\ \hline
1 & none & none \\
2 & t80 & none \\
3 & t72 & none  
\end{tabular}
\qquad\qquad
\begin{tabular}{r|cc}
\textbf{State} & \textbf{Vehicle} & \textbf{Location} \\ \hline
4 & none & Loc2 \\
5 & t80 & Loc2 \\
6 & t72 & Loc2 \\
\end{tabular}
\end{center}

The object ``t70'' in the first random variable has a null probability, and
hence it does not appear in any compatible agent state.
In states 1--3, the location is unknown.
In states 1 and 4, the vehicle in the image is unknown.
\end{Example}

We use the notation $\cmpos{\pagstate}$ to denote the set of all
ordinary agent states that are compatible with a $\pagstate$.
We now define the notion of a probabilistic Kripke structure.

\begin{definition}[Probabilistic Kripke Structure]\mbox{}\\
A \emph{probabilistic  Kripke structure} is a pair
$(\cals,\wp)$ where $\cals$ is a set of ordinary states,
and $\wp: \cals \to [0,1]$ is a mapping such that
$\sum_{\agstate\in\cals} \wp(\agstate) = 1$.
\end{definition}

\begin{definition}[Compatible
Probabilistic Kripke Structure]
Let  $\pagstate$ be a probabilistic agent state.   A coherent probabilistic
Kripke structure $(\cmpos{\pagstate},\wp)$ is said to be \emph{compatible}
with $\pagstate$ iff
for every ground code call $\cc{\aga}{f}{d_1,\ldots ,d_n}$, for every
random variable $(X,\wp')\in\evl{\cc{\aga}{f}{d_1,\ldots
,d_n}}{\pagstate}$, and for each object  $o$, it is the case that:
\[\sum_{o\in\evl{\cc{\aga}{f}{d_1,\ldots ,d_n}}{\agstate}
}
\wp(\agstate) =
\left\{\begin{array}{ll}
\wp'(o) & \text{ if } o\in X \\
0 & otherwise
\end{array}\right.
\]
\end{definition}
By definition, two distinct objects from the same 
random variable
cannot appear in the same compatible state.
If such a random variable has a complete probability distribution, then
in any compatible Kripke structure, the sum of the probabilities of the
states  containing one of its objects is equal to 1.
This means that any (compatible) agent state containing no such
objects will have a null probability, avoiding the intuitive
inconsistency pointed in example~\ref{example_compatible_state}.

\begin{Example}
\label{example_compatible_kripke}
Considering the situation in example~\ref{example_compatible_state} on
the previous page, a probabilistic Kripke structure compatible with $\pagstate$ is
$\langle \cmpos{\pagstate}, \wp\rangle $, with the following probability distribution:
\begin{center}
\begin{tabular}{r|cc}
\textbf{State} & \textbf{Probability} \\ \hline
1 & 0 \\
2 & 0.1 \\
3 & 0.1  
\end{tabular}
\qquad\qquad
\begin{tabular}{r|cc}
\textbf{State} & \textbf{Probability} \\ \hline
4 & 0 \\
5 & 0.2 \\
6 & 0.6 \\
\end{tabular}
\end{center}

\end{Example}

The following result says that compatible Kripke structures always
exist.
\begin{theorem}[Existence of Compatible Probabilistic Kripke Structure]
\label{prop_comp_kripke}
Suppose $\pagstate$ is a probabilistic agent state.  Then there
is at least one
probabilistic
Kripke structure which is compatible with it.
\end{theorem}
Hence, given a probabilistic agent state $\pagstate$, we use
the notation $\cpks{\pagstate}$ to denote the set of
all probabilistic Kripke structures compatible with
$\pagstate$---this set is guaranteed to be nonempty by
the preceding result.  However, in almost all cases, 
$\cpks{\pagstate}$  contains an infinite number of elements.

\begin{theorem}[Existence of Infinitely Many  Kripke Structures]
\label{prop-atleast}
If a probabilistic state $\pagstate$ contains at least two random
variables (returned by the same code call or by two distinct ones)
containing at least one object with associated probability $0<p<1$,
then there exist an infinite number of probabilistic
Kripke structures compatible with $\pagstate$.
\end{theorem}
We are now in a position to specify what it means to execute
an action in a probabilistic Kripke structure.

\begin{definition}[Action Execution]
Suppose $\pagstate$ is a probabilistic
agent state.
A ground action $\alpha$ is said to be \emph{possibly executable
in $\pagstate$} iff
there is at least one probabilistic Kripke structure
$(\cmpos{\pagstate},\wp)\in \cpks{\pagstate}$, and an
ordinary agent state $\agstate$ in
$\cmpos{\pagstate}$ in which the precondition of $\alpha$
is true such that $\wp(\agstate) > 0$.
In this case, $\agstate$ \emph{witnesses} the executability
of $\alpha$.

 We say that $\alpha$ is \emph{executable with probability {\sf p}}
 in $\pagstate$ \iffdef
 \begin{eqnarray*}
 {\sf p} & = & \mbox{min}\{\wp(\agstate)\: |\: \agstate\in\cmpos{\pagstate}\:
 \mbox{witnesses the executability of}\: \alpha\}.
\end{eqnarray*}
\end{definition}
The following example illustrates this definition.

\begin{Example}
Let us consider the probabilistic agent state of
examples~\ref{example_compatible_state}
and~\ref{example_compatible_kripke} and the following actions:
\[\begin{array}{lll}
\alpha_1: & \pre{\alpha_1} = &
\IN{t70}{\cc{surv}{identify}{image1}} \\
\alpha_2: & \pre{\alpha_2} = &
\IN{\var{X}}{\cc{surv}{location}{image1}} \:\&\: \var{X} \not= Loc2 \\ 
\alpha_3: & \pre{\alpha_3} = & \IN{Loc2}{\cc{surv}{location}{image1}}\:
\& \\
& & \IN{t80}{\cc{surv}{identify}{image1}}
\end{array}\]
As stated before, the object ``t70'' cannot appear in any compatible 
agent state, and hence the action $\alpha_1$ is not possibly executable.
On the other hand, the precondition of $\alpha_2$ requires the
existence of an object other than ``Loc2'', which is known to be the
only one possibly returned by the corresponding code call, and hence
$\alpha_2$ is not possibly executable either. 
Eventually, the precondition of $\alpha_3$ requires the presence of
both objects ``Loc2'' and ``t80'', which is true in the agent
state number 5 described in the above examples. As this state has a non null
probability, $\alpha_3$ is possibly executable and the agent state
witnesses its executability. 
\end{Example}

We are now ready to define the new probabilistic Kripke
structure that results when an action is executed in it.

\begin{definition}[Result of Action $(\theta,\gamma)$-Execution]
\label{def_result_action_execution}
Suppose  $\pagstate$ is a probabilistic agent state,
$\alpha(\vecx)$ is an action and $\gamma$ is
a ground substitution for
all variables occurring in the precondition, add, and delete list of
$\alpha(\vecx)\theta$.
Let $(\cmpos{\pagstate},\wp)$ be a probabilistic Kripke structure
that contains a witness $\agstate$ to the possible executability of
$\alpha(\vecx)\theta$.
The \emph{result of executing} action $\alpha(\vecx)$
under substitutions $\theta,\gamma$ in probabilistic Kripke structure
$\pks{\pagstate}=(\cals,\wp)$ is a new Kripke structure
$(\cals',\wp')$ defined in the following way:

\begin{enumerate}

\item $\cals' = \{
\textrm{map}_{\alpha(\vecx),\theta,\gamma}(\agstate) \: | \:
\agstate \in \cals \}$
where $\textrm{map}$ is defined as follows:
\[
\textrm{map}_{\alpha(\vecx),\theta,\gamma}(\agstate) =
\left\{\begin{array}{ll}
\textrm{apply}(\alpha(\vecx),\theta,\gamma,\agstate) & \text{
if } \agstate \in W \\
\agstate & \text{ otherwise}
\end{array}\right.
\]

\item $\wp'$ is defined as follows:
\[
\wp'(\agstate') = \sum \{ \wp(\agstate) \:|\: \agstate' =
\textrm{map}_{\alpha(\vecx),\theta,\gamma}(\agstate)\}
\]

\end{enumerate}

In the above definitions,  $W$ is the set of all witnesses in $\cals$ to the
executability of $\alpha(\vecx)\theta\gamma$.

The \emph{result of executing} action $\alpha(\vecx)$
under substitutions $\theta,\gamma$ in a set $\pksset$ of probabilistic
Kripke structures is $\{ \cals'\: |\: \cals'$ is obtained by executing
$\alpha(\vecx)$
under substitutions $\theta,\gamma$ on $\cals\in\pksset\}$.
\end{definition}
The definition causes the agent states in which $\alpha$'s
precondition is true to change, while those in which $\alpha$'s
precondition is false stay unchanged.  
The probability of each final state is the sum of the
probabilities of the corresponding (old) states.  This is
illustrated by the following example.

\begin{Example}
Let us consider the compatible probabilistic Kripke structure in
Example~\ref{example_compatible_kripke}, and the action $erase(X)$: 
\begin{description}
\item[Pre:] $\IN{\var{X}}{\cc{surv}{identify}{image1}}$ 
\item[Del:] $\IN{\var{X}}{\cc{surv}{identify}{image1}}$
\item[Add:] $\emptyset$
\end{description}
The result of executing action $erase(X)$ under substitutions
$\{X/t80\}, \epsilon$, is the probabilistic Kripke
structure $\langle\cals', \wp'\rangle $, briefly described by the following table:

\begin{center}
\begin{tabular}{r|cc|c}
\textbf{State} & \textbf{Vehicle} & \textbf{Location} & \textbf{Probability} \\ \hline
a & none & none & 0.1 \\
b & t72 & none & 0.1 \\
c & none & Loc2 & 0.2 \\
d & t72 & Loc2 & 0.6
\end{tabular}
\end{center}

\noindent i.e., the states 1 and 2 merge together yielding the new state ``a''
and their probabilities are summed. Similarly, states 4 and 5
yield  the new state ``c''. 
\end{Example}

The following result states that our definitions are coherent.

\begin{proposition}[Closure of Probabilistic Kripke Structures]
The result of $(\theta,\gamma)$-execution of an action in
a probabilistic Kripke structure is also a
probabilistic Kripke structure.
\end{proposition}
\begin{proof}
Let $\langle\cals, \wp\rangle $ be the original Kripke structure, and $\langle\cals',
\wp'\rangle $ the result of executing the action.
We just need to show that
$\sum\{\wp'(\agstate')\:|\:\agstate'\in\cals'\} = 1$.
Using Definition~\vref{def_result_action_execution}:
\[
\sum_{\agstate'\in\cals'} \wp'(\agstate') = \sum_{\agstate'\in\cals'}
\sum_{\agstate' = \textrm{map}(\agstate)} \wp(\agstate) 
= \sum_{\agstate\in\cals} \wp(\agstate) = 1
\]
\end{proof}

\section{p-Feasible Status Sets}\label{p-feasible}
Probabilistic Feasible Status Sets prevent an action from being
executed unless its precondition is known for sure to be true (which
is exactly the intuitive reading of $\pagstate\statemodels{1}{1}
Pre(\alpha)$, for a probabilistic agent state $\pagstate$ and an action
$\alpha$).
Analogously, an action constraint has to be checked only if its
precondition is certainly true.
Finally, state consistency requires that the execution of the actions
does not corrupt the consistency of the original agent state, i.e. it
has to lead to an agent state where the integrity constraints are
with 100\% probability,

In this section, we define the concept of $p$-feasibility, where
$p$ is a probability. $p$-feasibility weakens the above requirements
to only requiring that preconditions are true with probability
$p$ (or higher).

\begin{definition}[Operator \pPApp]
Suppose $\pagprog$ is a probabilistic agent program, $\pagstate$
is a probabilistic agent state, and $\pross$ is a probabilistic status set.  Then
\[\begin{array}{ll} \pPApp = \{ \op\,\alpha\:| & \op\,\alpha \text{ is the head
    of
    a ground instance of a rule}\\
  & \text{$r$ in $\pagprog$ and:}
\end{array}\]

\begin{enumerate}
\item $B^+_{as}(r) \subseteq \pross$ and $\lnot.B^-_{as}(r) \cap \pross = 
\emptyset$;
\item For every annotated code call condition
  $\accc{\chi}{\ann{\ai_1}{\ai_2}}{\cstrat}$ in the body of $r$, it is
  the case that $\pagstate\statemodels{\ai_1}{\ai_2}
  \accc{\chi}{\ann{\ai_1}{\ai_2}}{\cstrat}$;
\item if $\op\in \{ \bfP, \bfO, \bfDo\}$, then the preconditions of
  $\alpha$ are true with probability $p$ or greater, i.e.
  $\pagstate\statemodels{p}{1} Pre(\alpha)$;
\item for every action status atom of the form $\op\, \beta$ in
  $B^+_{as}(r)$ such that $\op\in \{ \bfP, \bfO, \bfDo\}$, the
  preconditions of $\beta$ are true with probability $p$ or greater, i.e.
  $\pagstate\statemodels{p}{1} Pre(\beta)$ ~$\}$
\end{enumerate}
\end{definition}
The only difference between this definition and that of $\PApp$ is that the
entailment $\statemodels{1}{1}$ is replaced by the more general
$\statemodels{p}{1}$.

\begin{definition}[Deontic and Action p-Consistency]
A probabilistic status set $\pross$ is  \emph{deontically
  p-consistent} with respect to a probabilistic agent state $\pagstate$ \iffdef
it
satisfies the following rules for any ground action $\alpha$:
    \begin{itemize}
      \item If $\bfO\alpha \in \pross$, then $\bfW\alpha \notin \pross$.
      \item If $\bfP\alpha \in \pross$, then $\bfF\alpha \notin \pross$.
\item If $\bfP\alpha \in \pross$, then the preconditions of $\alpha$
  are true with probability $p$ or greater,
  i.e. $\pagstate\statemodels{p}{1} Pre(\alpha)$.
\end{itemize}
A  probabilistic status set $\pross$ is  \emph{action p-consistent} with
respect to an agent state $\pagstate$ iff for every action constraint
of the form
\begin{equation}
\label{c9-AC2}
\{ \alpha_1(\vecx_1),\ldots ,\alpha_k(\vecx_k)\} \hookleftarrow \chi
\end{equation}
either $\pagstate\not\statemodels{p}{1}\chi$ or
$\{ \alpha_1(\vecx_1),\ldots ,\alpha_k(\vecx_k)\}
\not\subseteq \pross$.
\end{definition}
Generalizing probabilistic state consistency to 
$p$-probabilistic state consistency may be done in two ways.

\begin{definition}[Probabilistic State p-Consistency]
A probabilistic status set $\pross$ is \emph{weakly
probabilistically state p-consistent} w.r.t.~state $\pagstate$ iff the
new state, ${\pagstate} ^ {'} =
\concur(\bfDo(\pross),\pagstate)$ obtained after
concurrently executing all actions of the form $\bfDo\alpha\in \pross$
satisfies all integrity constraints with probability greater than or equal
to $p$, i.e. for every integrity constraint $\psi \Rightarrow \chi$
either ${\pagstate }^ {'}\not\statemodels{p}{1} \psi $ or
${\pagstate}^ {'} \statemodels{p}{1} \chi $.
\\
We say that a probabilistic status set $\pross$ is
\emph{strongly probabilistically state p-consistent} w.r.t.~state $\pagstate$
iff the new state ${\pagstate}^ {'}$ satisfies the following
condition: if  $\pagstate$ satisfies the integrity constraints with
probability $\geq q$ ($q\in[0,1]$) then also ${\pagstate}^ {'}$ does so.
\end{definition}
These definitions induce two types of feasibility for arbitrary
probabilities $p$.
\begin{definition}[Weak (resp. Strong) p-Feasibility]
Let $\pagprog$ be an agent program and
  let $\pagstate$ be an agent state. Then, a probabilistic
  status set $\pross$ is a \emph{p-feasible probabilistic status set}
  for $\pagprog$ on $\pagstate$, if the following conditions
  hold:
\begin{description}
\item[(p-\pross1):]  $\pPApp \subseteq \pross$ (closure under the
  program rules);
\item[(p-\pross2):] $\pross$ is deontically
  and action p-consistent (deontic and action p-consistency);
\item[(p-\pross3):] $\pross$ is action closed
  and deontically closed (deontic and action closure);
\item[(p-\pross4):] $\pross$ is weakly (resp. strong) state p-consistent (state
p-consistency).
\end{description}
\end{definition}

\begin{remark}
If $S$ is a p-feasible probabilistic status set for $\pagprog$ on
$\pagstate$ and $0\leq q \leq p$, then $S$ \emph{is not always} q-feasible.
Indeed, $[q,1]\supseteq [p,1]$, and then for any formula $\phi$
$\pagstate\statemodels{p}{1}\phi$ implies
that $\pagstate\statemodels{q}{1}\phi$ (and analogously for
$\not\statemodels{p}{1}$ and $\not\statemodels{q}{1}$).
This means that all preconditions of actions, preconditions of action
constraints and integrity constraints which are verified for $p$ are
also verified for $q$. \\
The problem is that $\pPApp$ is anti-monotonic w.r.t. $p$,
as a smaller value for $p$ may allow a larger set of rules to be firable.
Then the closure under the program rules is not guaranteed any
more.
\end{remark}
The following example illustrates this point.

\begin{example}
Consider the following trivial program:
\[
\bfDo \alpha \gets.
\]
where $\pre{\alpha} = \IN{a}{\cc{d}{f}{}}$, and
$\evl{\cc{d}{f}{}}{\pagstate} = \{\langle\{a\}, \langle0.7\rangle \rangle \}$.
Suppose $\pross = \emptyset$. Conditions p-\pross 2--4 are true for any
value of $p$.  Note that 
0.8-$\PApp = \emptyset$ nd hence, \pross is 0.8-feasible. In contrast, 
we see that 0.6-$\PApp = \{ \bfDo\alpha\} \not\subseteq \pross$, and
hence \pross is not 0.6-feasible.
\end{example}

We can easily see that probabilistic feasibility is a particular case
p-feasibility:
\begin{proposition}
Let $\pagprog$ be a probabilistic agent program and
  let $\pagstate$ be a consistent (or equivalently 1-consistent) agent state. Then, a probabilistic
  status set $\pross$ is a feasible probabilistic status set \iff
  it is weakly 1-feasible \iff it is strongly 1-feasible.
\end{proposition}
\begin{proof}
All definitions for weak p-feasibility trivially coincide with those
for feasibility if p=1.
The distinction between weak and strong p-feasibility is just in the
definition of p-consistency, and we can easily see that for p=1 they
coincide, since a probability \emph{greater or equal to} 1 cannot be
but equal to 1.
\end{proof}
\subsection{p-Rational and p-Reasonable Status Sets}
The notions of Rational and Reasonable Status Sets can be
straightforwardly extended to those of p-Rational and p-Reasonable
status sets.

\begin{definition}[p-Rational Status Set]
A probabilistic status set $\pross$ is a \emph{p-rational probabilistic
status set}, if $\pross$ is a p-feasible probabilistic status set and
there exists no probabilistic status set $\pross' \subset \pross$
satisfying conditions $(p-\pross 1)$--$(p-\pross 3)$ of a p-feasible
probabilistic status set.
\end{definition}

Obviously, in the case that $\intcons=\emptyset$ (i.e., there are no
integrity constraints) p-rational status sets are simply
inclusion-minimal feasible status sets.

\begin{definition}[p-Reasonable Probabilistic Status Set]
  Let $\pagprog$ be a probabilistic agent program, let
  $\agstate_\scode$ be an agent state, and let $\pross$ be a
  probabilistic status set.

\begin{enumerate}
\item If $\pagprog$ is a positive probabilistic agent program, then
  $\pross$ is a \emph{p-reasonable probabilistic status set} for $\pagprog$ on
  $\agstate_\scode$, \iffdef $\pross$ is a p-rational probabilistic status set
for  $\pagprog$ on $\agstate_\scode$.
\item Exploiting the definition of
$red^\pross(\pagprog,\agstate_\scode)$ (see Definition
~\vref{reasonable PSS}), $\pross$ is a \emph{p-reasonable probabilistic status 
set} for
$\pagprog$ w.r.t.~$\agstate_\scode$, if it is a p-reasonable
probabilistic status set of the program
$red^\pross(\pagprog,\agstate_\scode)$ with respect to
$\agstate_\scode$. {}
\end{enumerate}
\end{definition}

It is easy to verify that all p-reasonable probabilistic status sets
are p-rational probabilistic status sets:

\begin{proposition}[p-Reasonable Status Sets are p-Rational]
Let $\pagprog$ be a probabilistic agent program and $\agstate_\scode$
an agent state. Then, every p-reasonable probabilistic status set  of $\pagprog$ on
$\agstate_\scode$ is a p-rational probabilistic status set  of $\pagprog$ on
$\agstate_\scode$.
\end{proposition}
\begin{proof}
Identical to the proof of Proposition ~\vref{c9-prop:reasonable-sub-rational}. 
\end{proof}
As in Section~\vref{c9-sec:computing}, we can define a fixpoint
operator and build an algorithm on its top to compute p-reasonable
status sets for positive programs.

\begin{definition}[Operator $\pSPP$]
    \[\pSPP(\pross) = \acl{\papp{\pagprog,\agstate_\scode}{\pross}}{},\]
\end{definition}

Operator $\pSPP$ can be computed by an algorithm identical to Algorithm~\vref{c9-alg1}, but for step 4, where the entailment
$\statemodels{1}{1}$ has to be replaced by $\statemodels{p}{1}$.
$\pSPP$ is monotonic and continuous and has a unique
least fixpoint.

\begin{lemma}[Monotonicity and Continuity of $\pSPP$]
Suppose $\pagprog$ is a positive \pap.  Then the operator $\pSPP$ is
monotone and continuous, i.e.
\begin{enumerate}
\item $\pross_1\subseteq \pross_2\Rightarrow p-\SPP(\pross_1)\subseteq p-\SPP(\pross_2)$,
\item $\pSPP^\omega$ is a fixpoint of $\pSPP$.  Moreover, it is the least
  fixpoint of $\pSPP$. (We assume the iterations of $\pSPP$ are
defined in the same way as the iterations of $\SPP$).
\end{enumerate}
\end{lemma}
The following result now follows immediately and has a proof
similar to that of Theorem~\vref{c9-theo:pos-lfp}.

\begin{theorem}[p-Rational Probab.~Status Sets as Least
  Fixpoints]
\label{p-rat-as-least}\mbox{}\\
Let $\pagprog$ be a positive probabilistic agent program, and let
$\agstate_\scode$ be an agent state.  Then, $\pross$ is a p-rational
probabilistic status set of $\pagprog$ on $\agstate_\scode$, \iff
$\pross = \lfp(\pSPP)$ and $\pross$ is a p-feasible probabilistic status
set.
\end{theorem}

Uniqueness of the p-reasonable status set (if it exists) holds too,
and then we can compute it by Algorithm~\vref{c9-alg2}, 
replacing--as usual---the entailment $\statemodels{1}{1}$ with
$\statemodels{p}{1}$.

Unfortunately, the resulting algorithm is not polynomial because
of the integrity constraint check in steps (11)--(12).  This will
be discussed in detail in Section~\ref{al-wic} below.  However, when
no integrity constraints are present, the algorithm is still polynomial.

\begin{theorem}[Polynomial Data Complexity]
  The problem of computing p-reasonable probabilistic
  status sets of positive \paps without Integrity Constraints (i.e.,
  $\intcons = \emptyset$) has polynomial data-complexity.
\end{theorem}

\subsection{Checking p-Consistency of Integrity Constraints}
\label{al-wic}
In this section, we provide an algorithm to check $p$-consistency
of an integrity constraint IC after an action have been executed in
state $\pagstate$, leading to a new state $\pagstate{'}$ assuming that
all integrity constraints are true in the original state
$\pagstate$. 
Suppose $O_1, \ldots, O_N$ are all states
compatible with ${\pagstate}$ while
$O'_1, \ldots, O'_{N'}$ are those compatible with the new
state ${\pagstate}^{'}$. Let $p_i,p_i'$ denote the probabilities
of $O_i,O'_i$ respectively.  Then consider the following
system of constraints:

\[
\begin{array}{l}
\mbox{\bf minimize $\Sigma_{O'_i\models IC} p'_i$ such that:} \\
\begin{array}{ll}
\mbox{\bf(K)} & \sum_{i=1}^N p_i = 1 \\
\mbox{\bf(CK)} & \forall i\in \{1,\ldots, E\}. \sum_{j: Obj_i\in O_j} p_j
= p(Obj_i) \\
\mbox{\bf(IC)} & \forall IC\in\intcons. 1 \geq \sum_{O_i: O_i\models IC} p_i
\geq p \\
\mbox{\bf(K $\to$ K')} & \forall i\in \{1, \ldots, N'\}. p'_i
= \sum_{O_j: O_j \stackrel{\alpha}{\longrightarrow} O'_i} p_j \\
%
%
\mbox{\bf(IG)} & \forall i\in\{1,\ldots,N\}. \max\{0, \sum_{Obj_k\in
O_i} p(Obj_k)+1-|O_i|\} \leq \\
& \leq p_i \leq \min_{Obj_k\in O_i} \{p(Obj_k)\}
%
%
\end{array}
\end{array}
\]
The objective function captures the probability of $IC$ being true in
the new Kripke structure.  \textbf{(K)} and \textbf{(CK)} define any
arbitrary compatible Kripke structure over $N$ states w.r.t.
$\pagstate$ (which cointains $E$ objects), \textbf{(IC)} expresses the
fact that our actual state has to be p-consistent, while
\textbf{(K$\pmb{\to}$ K')} defines the Kripke structure obtained after
the execution of action $\alpha$.
%
%
Eventually, \textbf{(IG)} gives an upper and a lower bound to the
probability of worlds (it extends the Bool expression for conjunction
of events of unknown inter-relation).
%
%
It is easy to see that a straightforward implementation of this
algorithm requires exponential time and space.

\section{Related Work}
There has been an incredible amount of work on uncertainty in
knowledge based and database systems \cite{shaf-peal-90}.  However, almost all
this work assumes that we are reasoning with logic or with Bayesian
nets \cite{koll-98} and most work proceeds under strong assumptions
about the relationships between events (e.g. most Bayesian approaches
assume conditional independence between events, while other approaches
such as \cite{fagi-etal-90,ng-subr-93} assume that we have no knowledge of the
dependencies between events).

This paper introduces techniques to allow an agent developer to
encode different assumptions about the relationships between events,
when writing probabilistic agent programs.  The idea of conjunction
strategies to facilitate this was first introduced in
the \textsf{ProbView} system \cite{laks-etal-97} in an attempt to allow
users querying probabilistic relational databases to express in their
query, their knowledge of the dependencies between events.  Later,
\cite{dekt-subra-97} extended the use of conjunction and disjunction strategies
to the case of logic programs.  In this paper, the idea of
conjunction  strategies are applied in the context of
deontic-logic based agent programs.  We are not aware of any extant
work on allowing flexible dependency assumptions in the context of
logics and actions.

Research on epistemic logic (e.g.,
\cite{morg-88,moore-85,krau-lehm-88}) enables reasoning about what
is known and is not known at a given time. However, epistemic logics
have not been used as a representation in decision making and in
automated planning systems, perhaps, because the richness of these
languages makes efficient reasoning very difficult.
In contrast, our framework has polynomial data complexity.

\citeN{halp-tutt-92} study the semantics of reasoning about distributed
systems when uncertainty is present.  They develop a logic where a
process has knowledge about the probability of events which
facilitates decision-making by the process.  We, on the other hand,
consider probabilistic states, and as argued in~\cite{dix-etal-00}
this also allows us to reason about probabilistic beliefs, i.e.
probabilities are assigned to the agents' beliefs about events, rather
than to the events themselves.  That is, in Halpern's work
\cite{halp-tutt-92}, the beliefs of the agent are CERTAIN, but in our
framework, the beliefs of the agent may themselves be uncertain (with
the phenomenon when they are certain being a special case of our
framework).

\citeN{poole-97} presented a framework that allows a natural
specification of multi-agent decision problems. It extends logic with a new
way to handle and think about non-determinism and uncertainty in terms
of independent choices made by various agents, including nature and a
logic program that gives the consequence of choices.  It has general
independence assumption.  This work is more expressive than ours, but
its generality leads to complexity problems and to difficulties in
using the framework.

\citeN{hadd-91} developed a logic that allows to write sentences
that describe uncertainty in the state of the world, uncertainty of
action effects, combine possibility and chance, distinguish between
truth and chance and express information about probability
distributions. He uses model theoretic semantics and demonstrates how
his logic can be used to specification various reasoning and planning
problems. The main purpose of the specification is to prove
correctness, and not for programming of agents.
\citeN{kush-etal-95} model uncertainty about the
true state of the world with a probability distribution over the state
space.  Actions have uncertain effects, and each of these effects is
also modeled with a probability distribution. They seek plans whose
probability of success exceeds the threshold. They describe
\textsf{BURIDAN}, an implemented algorithm for probabilistic planning.
In contrast, we focus on programming agents, rather than on how
agents will construct plans.  Other
researchers extended Kushmerick et al.'s model to increase the
efficiency of the planning \cite{hadd-etal-96} or to more realistic
domains \cite{doan-96}.  \citeN{thie-etal-95} developed a
framework for anytime generation of plans under incomplete and
ambiguous knowledge and actions with alternative and context dependent
effects.

\citeN{kael-etal-98} propose using
\emph{partially observable Markov decisionprocesses} (\textsf{POMDP}s) for planning
under uncertainty.  Similar to \textsf{BURIDAN} they use a probability
distributions over states to express uncertainty about the situation
of the agent. They also consider the problem of non-deterministic
actions and getting feedback from the environment which we
mentioned only briefly.

\section{Conclusions}
Agents are programs that autonomously react to changes in their
environment by taking appropriate actions.  In \cite{eite-etal-99a},
the authors have proposed a framework within which agents may be
built on top of an existing body of legacy code, and/or on top
of specialized data structures appropriate for the intended functionality
of the agent being built.

However, there are an increasing number of applications where
agents are uncertain about what is true in their state (or environment).
Such situations occur all the time in image identification programs,
in programs that predict future events (such as battlefield events,
stock market events, etc.), and in scenarios where an agent $\aga$
attempts to predict what an agent $\agb$ will do.

In this paper, we first introduce the concept of a probabilistic
code call, which is a mechanism to describe uncertain application
program interfaces for arbitrary functions over arbitrary data
types.  Based on this concept, we define probabilistic agent programs
--- sets of rules that encode the operating principles of an agent.
Such rules encode the
\emph{probabilistic} conditions under which an agent is obliged
to take some actions, permitted to take some actions and/or forbidden
to take some actions.

We then provide two broad classes of semantics for such ``probabilistic
agents.''  In the first class of semantics, actions that are permitted,
obligatory or done, must have preconditions that are true with 100\%
probability in the
current agent state. 
In the second class of semantics (which use probabilistic variants
of Kripke structures), the actions that are  permitted,
obligatory or done, must have preconditions that are true with at least
a given probability. This latter class of semantics allows reasoning
by cases.
We provide complexity arguments showing that though the second
family of semantics is perhaps epistemologically more appealing than
the first, the second family of semantics is also computationally
more complex.

Finally, the paper includes algorithms to compute the semantics of
probabilistic agent programs, as long as such programs are negation
free.

Future work on probabilistic agent programs will focus on computing
the semantics of {\pap}s that contain negation.  A detailed study
of computational complexity is also envisaged.  We are also interested
in identifying polynomially computable fragments of {\pap}s and
implementing them on top of the current \impact implementation.  Last,
but not least, \impact has been used in a battlefield
monitoring application where there is considerable uncertainty
in predicting tactical enemy movements.  We hope to build an
application of {\pap}s addressing this problem, once the
implementation of {\pap}s is complete.


\appendix

\section{Agent Programs without Uncertainty}\label{app-def}

The following definitions are taken from \cite{eite-etal-99a}.
\subsection{Feasible, Rational and Reasonable Semantics}

\begin{definition}[Status Set]\label{c7-SS}
A \emph{status set} is any set $S$ of
ground action status atoms over $\scode$.  For any operator $\textsf{Op} \in
\{ \bfP,\bfDo,\bfF,\bfO,\bfW \}$,
we denote by $\textsf{Op}(S)$ the set $\textsf{Op}(S) = \{ \alpha \mid 
\textsf{Op}(\alpha)
 \in S\}$.
\end{definition}

\begin{definition}[Deontic and Action Consistency]
\label{c6-def:dac}
A status set $S$ is called \emph{deontically consistent}, \iffdef it
satisfies the following rules for any ground action $\alpha$:
    \begin{itemize}
      \item If $\bfO\alpha \in S$, then $\bfW\alpha \notin S$
      \item If $\bfP\alpha \in S$, then $\bfF\alpha \notin S$
\item If $\bfP\alpha \in S$, then $\agstate_\scode \models
\exists^*\pre{\alpha}$, where $\exists^*\pre{\alpha}$ denotes the
existential closure of $\pre{\alpha}$, i.e., all free variables in
$\pre{\alpha}$ are governed by an existential quantifier.  This
condition means that the action $\alpha$ is in fact executable in the
state $\agstate_\scode$.

\end{itemize}
A status set $S$ is called \emph{action consistent}, if
$S,\agstate_\scode \models \actcons$ holds.
\end{definition}

Besides consistency, we also wish that the presence of certain atoms in
$S$ entails the presence of other atoms in $S$. For example, if
$\bfO\alpha$ is in $S$, then we expect that $\bfP\alpha$ is also in
$S$, and if $\bfO\alpha$ is in $S$, then we would like to have
$\bfDo\alpha$ in $S$. This is captured by the concept of deontic and
action closure.

\begin{definition}[Deontic and Action Closure]
\label{c6-def:dcl}
The \emph{deontic closure} of a status $S$, denoted
$\dcl{S}{}$,  is the closure of $S$ under the
rule
\begin{itemize}
    \item[] {\it If $\bfO\alpha \in S$, then $\bfP\alpha \in S$}
\end{itemize}
where $\alpha$ is any ground action. We say that $S$ is \emph{deontically
closed}, if $S = \dcl{S}{}$ holds.

The \emph{action closure} of a status set $S$, denoted $\acl{S}{}$, is
the closure of $S$ under the rules
\begin{itemize}
    \item[] {\it If $\bfO\alpha \in S$, then $\bfDo\alpha \in S$}
    \item[] {\it If $\bfDo\alpha \in S$, then  $\bfP\alpha \in S$}
\end{itemize}
where $\alpha$ is any ground action.
We say that a status $S$ is action-closed, if $S = \acl{S}{}$ holds.
\end{definition}
The following straightforward results shows
that status sets that are action-closed
are also deontically closed, i.e.

\begin{definition}[Operator \App]
\label{c7-def:appold}
Suppose $\agprog$ is an agent program, and $\agstate_\scode$ is an
agent state. Then, \App is defined to be the set of all ground action
status atoms $A$ such that there exists a rule in $P$ having a ground
instance of the form $r:$ $A \la L_1,\ldots, L_n$ such that

\begin{enumerate}
\item $B^+_{as}(r) \subseteq S$ and $\lnot.B^-_{as}(r) \cap S =
  \emptyset$, and
\item every code call $\chi \in B^+_{cc}(r)$ succeeds in
  $\agstate_\scode$, and
\item every code call $\chi \in \lnot.B^-_{cc}(r)$ does not succeed in
  $\agstate_\scode$, and
\item for every atom $\textsf{Op}(\alpha) \in B^+(r) \cup \{ A\}$ such that $\textsf{Op}
  \in \{ \bfP, \bfO, \bfDo\}$, the action $\alpha$ is executable in
  state $\agstate_\scode$. {}
\end{enumerate}
\end{definition}

Note that part (4) of the above definition only applies to the
``positive'' modes $\bfP,\bfO,\bfDo$.  It does not apply to atoms of
the form $\bfF\alpha$ as such actions are not executed, nor does it
apply to atoms of the form $\bfW\alpha$, because execution of an
action might be (vacuously) waived, if its prerequisites are not
fulfilled.

Our approach is to base the semantics of agent programs on consistent
and closed status sets. However, we have to take into account the
rules of the program as well as integrity constraints. This leads us
to the notion of a feasible status set.

\begin{definition}[Feasible Status Set]\label{c6-feasSS}
Let $\agprog$ be an agent program and let $\agstate_\scode$ be an agent
state. Then, a status set $S$ is a \emph{feasible status set} for $\agprog$
on $\agstate_\scode$, if the following conditions hold:

\begin{description}
\item[(S1):] (closure under the program rules) \quad
  $\App \subseteq S$;
\item[(S2)] (deontic and action consistency) \quad $S$ is deontically
  and action consistent;
\item[(S3)] (deontic and action closure) \quad $S$ is action closed
  and deontically closed;
\item[(S4)] (state consistency) \quad $\agstate'_\scode \models \intcons$,
  where $\agstate'_\scode = \textrm{apply}(\bfDo(S),\agstate_\scode)$
  is the state which results after taking all actions in $\bfDo(S)$ on
  the state $\agstate_\scode$.
\end{description}
\end{definition}

\begin{definition}[Groundedness; Rational Status Set]
  A status set $S$ is \emph{grounded}, if there exists no status set
  $S'\neq S$ such that $S' \subseteq S$ and $S'$ satisfies
  conditions $(S1)$--$(S3)$ of a feasible status set.

A status set $S$ is a \emph{rational status set}, if
$S$ is a feasible status set and $S$ is grounded.
\end{definition}

\begin{definition}[Reasonable Status Set]
  Let $\agprog$ be an agent program, let $\agstate_\scode$ be an agent
  state, and let $S$ be a status set.

\begin{enumerate}
\item If $\agprog$ is a positive agent program, then $S$ is a
  \emph{reasonable status set} for $\agprog$ on $\agstate_\scode$,
  if and only if $S$ is a rational status set for $\agprog$ on
  $\agstate_\scode$.
\item
The reduct of $\agprog$ w.r.t.\ $S$ and $\agstate_\scode$, denoted by
$red^S(\agprog,\agstate_\scode)$, is the program which is obtained from the ground
instances of the rules in $\agprog$ over $\agstate_\scode$ as follows.
\begin{enumerate}
\item First, remove every rule $r$ such that $B^-_{as}(r) \cap S \neq
  \emptyset$;
\item Remove all atoms in $B^-_{as}(r)$ from the remaining rules.
\end{enumerate}
Then $S$ is a \emph{reasonable status set} for $\agprog$ w.r.t.\
$\agstate_\scode$, if it is a reasonable status set of the program
$red^S(\agprog,\agstate_\scode)$ with respect to $\agstate_\scode$. {}
\end{enumerate}
\end{definition}

\section{Proofs of Theorems}\label{app-proofs}
\begin{proof}\textbf{(of Proposition~\ref{c9-prop:feas})}\\
\begin{enumerate}
\item Suppose $\bfDo\alpha\in \pross$. Then, as $\pross$ is feasible, we
  know that $\pross=\acl{\pross}{}$, and hence $\bfP\alpha\in \pross$.  As $\pross$ is
  feasible, and hence deontically consistent, the third condition of
  deontic consistency specifies that $\alpha$'s precondition is true
  in state $\agstate_\scode$.
  
\item This follows immediately because as $\pross$ is feasible, we have
  $\pross=\acl{\pross}{}$. The second condition defining $\acl{\pross}{}$, when
  written in contrapositive form, states that $\bfP\alpha\notin \pross$
  implies that $\bfDo\alpha\notin \pross$.
  
\item As $\pross$ is feasible, $\pross=\acl{\pross}{}$. The first condition specifying
  $\acl{\pross}{}$ allows us to infer that $\bfO\alpha\in \pross$ implies that
  $\bfDo\alpha\in \pross$.  The result follows immediately from part (1) of
  this proposition.
  
\item From the above argument, as $\pross=\acl{\pross}{}$, we can conclude that
  $\bfO\alpha\in \pross$ implies that $\bfP\alpha\in \pross$.  By the deontic
  consistency requirement, $\bfF\alpha\notin \pross$.
\end{enumerate}
\end{proof}

\begin{proof}\textbf{(of Theorem~\ref{c9-prop:reasonable-sub-rational})}\\
 In order to show that a reasonable probabilistic status set\index{status set!probabilistic} $\pross$ of
  $\pagprog$ is a rational status of $\pagprog$, we have to verify (1)
  that $\pross$ is a feasible probabilistic status set\index{status set!probabilistic} and (2) that $\pross$ is grounded.
  
  Since $\pross$ is a reasonable probabilistic status set\index{status set!probabilistic} of $\pagprog$, it is a
  rational probabilistic status set\index{status set!probabilistic} of $\pagprog' =
  red^\pross(\pagprog,\agstate_\scode)$, i.e., a feasible and grounded probabilistic
   status set of $\pagprog'$.  Since the conditions
   $(\pross2)$--$(\pross4)$ of the definition of feasible
   probabilistic status set\index{status set!probabilistic} depend
  only on $\pross$ and $\agstate_\scode$ but not on the program, this
  means that for showing (1) it remains to check that $(\pross1)$
  (closure under the program rules) is satisfied.
  
  Let thus $r$ be a ground instance of a rule from $\pagprog$. Suppose
  the body $B(r)$ of $r$ satisfies the conditions 1.--4.\ of $(\pross
  1)$. Then, by the definition of
  $red^\pross(\pagprog,\agstate_\scode)$, we have that the reduct of
  the rule $r$, obtained by removing all literals of $B^-_{as}(r)$
  from the body, is in $\pagprog'$. Since $\pross$ is closed under the
  rules of $\pagprog'$, we have $H(r) \in \pross$. Thus, $\pross$ is
  closed under the rules of $\pagprog$, and hence $(\pross1)$ is
  satisfied. As a consequence, (1) holds.
  
  For (2), we suppose $\pross$ is not grounded, i.e., that some
  smaller $\pross ' \subset \pross$ satisfies $(\pross1)$--$(\pross3)$
  for $\pagprog$, and derive a contradiction. If $\pross'$ satisfies
  $(\pross1)$ for $\pagprog$, then $\pross'$ satisfies $(\pross1)$
  for $\pagprog '$. For, if $r$ is a rule from $\pagprog '$ such that
  1.--4.\ of $(\pross1)$ hold for $\pross'$, then there is a ground
  rule $r'$ of $\pagprog$ such that $r$ is obtained from $r'$ in the
  construction of $red^\pross(\pagprog,\agstate_\scode)$ and, as
  easily seen, 1.--4. of $(\pross1)$ hold for $\pross'$. Since
  $\pross'$ satisfies $(\pross1)$ for $\pagprog$, we have $H(r) \in
  \pross'$. It follows that $\pross'$ satisfies $(\pross1)$ for
  $\pagprog'$. Furthermore, since $(\pross2)$ and $(\pross3)$ do no
  depend on the program, also $(\pross2)$ and $(\pross3)$ are
  satisfied for $\pross'$ w.r.t.\ $\pagprog'$. This means that
  $\pross$ is not a rational probabilistic status set\index{status set!probabilistic} of $\pagprog'$, which is the
  desired contradiction.

Thus, (1) and (2) hold, which proves the result.
\end{proof}

\begin{proof}\textbf{(of Theorem \ref{c9-prop:pos-rat-exists})}\\
By definition of rationality, we know that if $\pross$ is a rational status
set of $\pagprog$ then it must be a feasible probabilistic status set\index{status set!probabilistic} as well. 

Suppose $\pagprog$ has a feasible probabilistic status set.  Then the set of all
feasible  probabilistic status sets of $\pagprog$ on $\agstate_\scode$ has a
non-empty set of inclusion-minimal elements.  Indeed, from the
grounding of the probabilistic agent program, we can remove all rules
which violate the conditions 2.-4.\ of the operator \PApp, and can
remove literals involving code calls from the remaining rules.
Moreover, the deontic and action closure conditions can be
incorporated into the program via rules.  Thus, we end up with a set
$T$ of propositional clauses, whose models are feasible  probabilistic 
status sets of
$\pagprog$. Since $\pagprog$ has a feasible  probabilistic status set, $T$ has a
model, i.e., an assignment to the propositional atoms which satisfies
all clauses in $T$. Now, each satisfiable set of clauses in a
countable language posseses at least one minimal model (w.r.t.\ 
inclusion, i.w., a $\subseteq$-minimal set of atoms is assigned the value
\true); this can be shown applying the same technique which proves
that every such set of clauses can be extended to a maximal
satisfiable set of clauses.  Thus, $T$ has at least one minimal model.
As easily seen, any such model is a minimal feasible  probabilistic status set\index{status set!probabilistic} of
$\pagprog$.

Suppose now $\pross'$ is one of the minimal feasible  probabilistic status sets of
$\pagprog$ on $\agstate_\scode$.  Then (as we show below) $\pross'$ is
grounded, and hence a rational  probabilistic status set.

To show that $\pross'$ is grounded, we need to show that $\pross'$ satisfies conditions
(\pross1)--(\pross3) of the definition of feasible  probabilistic status set---this is true because
$\pross'$ is feasible. In addition, we need to show that no strict subset
$\pross^\star$ of $\pross$ satisfies conditions (\pross1)--(\pross3).

Suppose there is a strict subset $\pross^\star$ of $\pross$ satisfying
conditions (\pross1)--(\pross3).  Then, as $\intcons=\emptyset$,
$\pross^\star$ also satisfies condition (\pross4) of the definition of
feasibility, and hence $\pross^\star$ is a feasible  probabilistic status set.  But this
contradicts the inclusion minimality of $\pross'$, and hence, we may
infer that $\pross'$ has no strict subset $\pross^\star$ of $\pross$
satisfying conditions (\pross1)--(\pross3).  Thus, $\pross'$ is
grounded, and we are done.
\end{proof}

\begin{proof}\textbf{(of  Theorem~\ref{prop_comp_kripke})}\\
For each random variable $V_i=(X_i, \wp_i)$ returned by some ground code call
condition in the probabilistic state $\pagstate$, let us define its {\em
normalized} version $V_i'=(X_i', \wp_i')$ where:
\[\begin{array}{l}
X_i' = \{ x\: |\: x\in X_i \text{ and } \wp_i(x)>0 \} \cup \{ \epsilon\: | \:
\sum_{x\in X_i} \wp_i(x) < 1 \}; \\
\wp_i'(x) = \left\{\begin{array}{ll} 
\wp_i(x) & \text{if } x\in X_i'\setminus\{\epsilon\} \\
1 - \sum_{x\in X_i} \wp_i(x) & \text{if } x = \epsilon \text{ and }
\epsilon \in X_i' 
\end{array}\right.
\end{array}\]
i.e., we delete the zero-probability elements and add the extra one
$\epsilon$ (which stands for ``none of the above'') whenever the
distribution $\wp_i$ is incomplete.
Now we can see that each tuple $\overline{x} = \langle x_1,\ldots,x_n\rangle $ in
the Cartesian product
$\overline{X'} = X_1'\times \cdots \times X_n'$ corresponds to a
distinct compatible
state $\agstate$ w.r.t. $\pagstate$. In $\agstate$, a ground code
call returns an object $o$ iff in the probabilistic state $\pagstate$
it returns a variable $V_i$ such that $x_i = o$.
Let associate to each state $\agstate$ of this kind the value 
\[
\wp^*(\agstate) = \wp_1'(x_1)\cdots\wp_n'(x_n)
\]
and set $\wp^*(\agstate) = 0$ for all the other states. We can easily
verify that $\langle \cmpos{\pagstate}, \wp^*\rangle $ is a compatible probabilistic
Kripke structure for $\pagstate$: for each random variable $V_i$
returned in state $\pagstate$ and each object $o\in X_i$ if
$\wp_i(o) = 0$ then $o\not\in X'$, so it could appear only in the
zero-probability states; otherwise:
\[
\sum_{o\in\agstate}\wp^*(\agstate) =
\sum_{\overline{x}\in\overline{X'}, x_i=o}
\prod_{j=0}^n \wp_j'(x_j) = 
\wp_i'(o)\sum_{\overline{x}\in\overline{X'}, x_i=o}
\prod_{j\not=i} \wp_j'(x_j) = \wp_i'(o) = \wp_i(o) 
\]
Both cases satisfy the condition for compatibility. Finally, it is
easy to verify that $\langle  \cmpos{\pagstate}, \wp^*\rangle $ is really a
probabilistic Kripke structure:
\[
\sum_{\agstate} \wp^*(\agstate) = \sum_{\overline{x}\in\overline{X'}}
\prod_{j=0}^n \wp_j'(x_j) = \sum_{x_1\in X_1'}\wp_1'(x_1) = 1
\]
\end{proof}
\begin{proof}\textbf{(of Theorem~\ref{prop-atleast})}\\
Let us consider the compatible Kripke structure described in the proof of
Proposition~\vref{prop_comp_kripke}, and let us assume that $V_1$ and
$V_2$ are the variables required in the thesis.
The corresponding {\em completed} versions $V'_1$ and $V'_2$ will then
contain at least two non zero-probability objects (one of them could
be the extra object $\epsilon$), respectively $a_1, b_1$ and $a_2,b_2
$.
Now let choose an arbitrary real number $\delta$ such that:
\[
0 \leq \delta \leq \min_{x\in\{a_1,b_1\}, y\in\{a_2,b_2\}}\{\wp'_1(x)
\wp'_2(y)\} 
\]
We can build a Kripke structure $\langle \cmpos{\pagstate},
\wp^\delta\rangle $, where $\wp^\delta$ is defined in the same way as $\wp^*$
but replacing $\wp'_1(x_1)\wp'_2(x_2)$ by $\phi(x_1,x_2)$, which in
turn is defined in the following way:
\[
\phi(x_1,x_2) = \left\{\begin{array}{ll}
\wp'_1(x_1)\wp'_2(x_2)-\delta & \text{ if }\langle x_1,x_2\rangle  \in \{\langle a_1,a_2\rangle ,
 \langle b_1,b_2\rangle \} \\
\wp'_1(x_1)\wp'_2(x_2)+\delta & \text{ if }\langle x_1,x_2\rangle  \in \{\langle a_1,b_2\rangle ,
 \langle b_1,a_2\rangle \} \\
\wp'_1(x_1)\wp'_2(x_2) & \text{ otherwise}
\end{array}\right.
\]
It is easy to verify that it is a compatible Kripke structure. Since
$\delta$ can be arbitrarily chosen within a non-point interval, we can
obtain an infinite number of distinct compatible Kripke structures.
\end{proof}

\begin{proof}\textbf{(of Theorem~\ref{p-rat-as-least})}\\
$(\Rightarrow)$ Suppose $\pross = \lfp(\SPP)$ a rational probabilistic status set\index{status set!probabilistic} of $\pagprog$
on $\agstate_\scode$. Then, $\pross$ is feasible by definition of rational
probabilistic status
set. By Lemma~\ref{c9-lem:fixpoint}, $\pross$ is a pre-fixpoint of
$\SPP$. Since $\SPP$ is monotone, it has by the Knaster-Tarski Theorem a
least pre-fixpoint, which coincides with $\lfp(\SPP)$ (see
\cite{apt-90,lloy-84}).  Thus, $\lfp(\SPP) \subseteq \pross$. Clearly,
$\lfp(\SPP)$ satisfies $(\pross1)$ and $(\pross3)$; moreover, $\lfp(\SPP)$ satisfies
$(\pross2)$, as $\pross$ satisfies $(\pross2)$ and this property is hereditary. By the
definition of rational probabilistic status set, it follows $\lfp(\SPP) = \pross$.

$(\Leftarrow)$ Suppose $\pross = \lfp(\SPP)$ is a feasible probabilistic status set. Since
every probabilistic status set\index{status set!probabilistic} $\pross'$ which satisfies $(\pross1)$--$(\pross3 )$ is a pre-fixpoint of
$\SPP$ and $\lfp(\SPP)$ is the least prefix point, $\pross '\subseteq \pross$
implies $\pross=\pross '$. It follows that $\pross$ is rational.

Notice that in case of positive programs, $\lfp(\SPP)$ always
satisfies the conditions $(\pross1)$ and $(\pross3)$ of a feasible
probabilistic status set\index{status set!probabilistic} (i.e., all
closure conditions), and thus is a rational probabilistic status
set\index{status set!probabilistic} if it satisfies $(\pross2)$ and
$(\pross4)$, i.e., the consistency criteria. The uniqueness of the
rational probabilistic status set\index{status set!probabilistic} is
immediate from the previous theorem.
\end{proof}
\end{document}